\documentclass[final,12pt]{clear2026} 
\usepackage{caption}
\usepackage{sidecap}

\usepackage{booktabs}
\usepackage{tikz} 
\usetikzlibrary{positioning,calc,arrows,arrows.meta,bending} 

\tikzset{
  obs/.style={circle, draw, thick, minimum size=18pt, inner sep=0pt, font=\small, fill=white},
  noise/.style={rectangle, draw, thick, rounded corners=2pt, minimum width=18pt, minimum height=12pt, inner sep=1pt, font=\scriptsize, fill=white},
  lat/.style={circle, draw, thick, dashed, minimum size=18pt, inner sep=0pt, font=\small, fill=white},
  dir/.style={-{Latex[length=2.4mm]}, line width=0.9pt}, 
  bidi/.style={<->, >=Latex, line width=0.9pt},         
  lbl/.style={font=\scriptsize, fill=white, inner sep=1pt}
}

\title[Causal Discovery with Mixed Confounding]{Causal Discovery with Mixed Latent Confounding via Precision Decomposition}
\usepackage{times}

\clearauthor{%
  \Name{Amir Asiaee} \Email{amir.asiaeetaheri@vumc.org} 
  \AND
  \Name{Samhita Pal} \Email{samhita.pal@vumc.org}\\
  \addr Department of Biostatistics\\
  Vanderbilt University Medical Center\\
  Nashville, TN 37232, USA
  \AND
  \Name{James O'quinn} \Email{oquinnjm@proton.me}\\
  \addr Department of Computer Science\\
  Johns Hopkins University\\
  Baltimore, MD 21218, USA
  \AND
  \Name{James P. Long} \Email{jplong@mdanderson.org}\\
  \addr Department of Biostatistics\\
  MD Anderson Cancer Center\\
  Houston, TX 77030, USA
}

\usepackage{bm} 

\usepackage{enumitem}

\newcommand{\rank}{\mathrm{rank}}

\makeatletter
\def\ps@jmlrtps{%
  \let\@mkboth\@gobbletwo
  \def\@oddhead{}%
  \let\@evenhead\@oddhead
  \def\@oddfoot{\@titlefoot}
  \let\@evenfoot\@oddfoot
}
\makeatother

\newcommand{\vct}[1]{\boldsymbol{#1}}     
\newcommand{\mat}[1]{\boldsymbol{#1}}     

\newcommand{\x}{\vct{x}}

\newcommand{\B}{\mat{B}}
\newcommand{\U}{\mat{U}}      
\newcommand{\V}{\mat{V}}      
\newcommand{\Lmat}{\mat{L}}   
\newcommand{\Smat}{\mat{S}}   

\newcommand{\The}{\mat{\Theta}}

\newcommand{\argmin}{\mathop{\mathrm{arg\,min}}}

\newtheorem{assumption}{Assumption}

\begin{document}

\makeatletter
\gdef\@editor{}%
\makeatother

\maketitle


\begin{abstract}
We study causal discovery from observational data in linear Gaussian systems affected by \emph{mixed latent confounding}, where some unobserved factors act broadly across many variables while others influence only small subsets. This setting is common in practice and poses a challenge for existing methods: differentiable and score-based DAG learners can misinterpret global latent effects as causal edges, while latent-variable graphical models recover only undirected structure. 

We propose \textsc{DCL-DECOR}, a modular, precision-led pipeline that separates these roles. The method first isolates pervasive latent effects by decomposing the observed precision matrix into a structured component and a low-rank component. The structured component corresponds to the conditional distribution after accounting for pervasive confounders and retains only local dependence induced by the causal graph and localized confounding. A correlated-noise DAG learner is then applied to this deconfounded representation to recover directed edges while modeling remaining structured error correlations, followed by a simple reconciliation step to enforce bow-freeness.

We provide identifiability results that characterize the recoverable causal target under mixed confounding and show how the overall problem reduces to well-studied subproblems with modular guarantees. Synthetic experiments that vary the strength and dimensionality of pervasive confounding demonstrate consistent improvements in directed edge recovery over applying correlated-noise DAG learning directly to the confounded data.
\end{abstract}



\section{Introduction}

Causal graphs are the stories we tell about systems we cannot intervene on. In the linear Gaussian setting, that story is often interrupted by latent confounders: unobserved variables that nudge many observables in concert (pervasive effects) or quietly perturb a few of them at a time (sparse, local effects). The result is familiar to practitioners: covariance that looks too global to be explained by a small set of edges, and precision patterns that look too structured to be pure noise. If we ignore these latent forces, causal discovery tends to latch onto spurious correlations or hedge with large equivalence classes \citep{spirtes2000causation}. If we over-correct, we erase real signal.

Two types of confounding can generally arise. Pervasive confounding is driven by a small number of latent factors that load broadly across many measured variables (e.g., batch effects, global cell state, or shared environmental signals), and this regime has been extensively studied through sparse-plus-low-rank / latent-variable graphical models and approximate factor models \citep{chandrasekaran2012latent, frot2019robust}. In contrast, sparse (localized) confounding induces correlation only among a relatively small subset of variable pairs (e.g., pathway-specific hidden regulators or unmeasured co-regulators that affect only a few proteins/genes at a time), and while it is naturally represented using mixed-graph / correlated-error formalisms, it has received comparatively less systematic attention in scalable score-based causal discovery \citep{pal2025decor}. Our paper tackles \emph{causal discovery with mixed confounding}: both \textit{pervasive} low-rank effects and \textit{sparse} low-rank effects co-exist. We work with linear Gaussian structural equation models (SEMs),
\(
\x = \B^\top \x + \vct{\varepsilon},
\)
and model the exogenous noise as a homoskedastic idiosyncratic term plus two latent components: a dense low-rank factor \(\U\vct{u}\) (few hidden causes that touch many nodes) and a column-sparse low-rank factor \(\V\vct{v}\) (many hidden causes, each touching a small subset). This hybrid regime is common in practice: batch or device drift (\(\U\)) co-exists with pathway/module-specific influences (\(\V\)) in biology, finance, and recommendation.

\paragraph{Why mixed confounding is hard.}
In linear Gaussian models, latent variables generally render the DAG unidentifiable from a single observational environment \citep{spirtes2000causation}. Classical methods that assume causal sufficiency (no hidden nodes) return only a Markov-equivalence class \citep{chickering2002optimal}. Non-Gaussian or nonlinear assumptions can sometimes restore identifiability \citep{shimizu2006lingam,hoyer2009nonlinear}, but our focus is the strictly Gaussian, linear case with both pervasive and sparse confounders. In this regime, undirected analogs have long exploited sparse+low-rank decompositions of precision or covariance to separate conditional structure from latent factors \citep{chandrasekaran2012latent}, yet turning those decompositions into \emph{directed} graphs that are robust to mixed confounding is non-trivial.

\paragraph{Our view: deconfound in precision, learn in data.}
We develop a modular pipeline that begins where latent-variable graphical modeling is strongest \emph{in the precision} and ends where modern continuous directed acyclic graph (DAG) learning thrives \emph{on per-sample residuals}. First, we decompose the observed precision \(\The\) into a sparse SPD matrix \(\Smat\) and a positive semidefinite low-rank matrix \(\Lmat\) (so \(\The \approx \Smat - \Lmat\)). At the population level, \(\Lmat\) captures the pervasive component, while \(\Smat\) encodes the DAG convolved with the remaining (sparse) confounding. We then invert \(\Smat\) to obtain a conditioned (pervasive-adjusted) covariance estimate, and apply the DECOR-GL algorithm \cite{pal2025decor} to separate the remaining confounding from directed effects and learn a weighted DAG.

\paragraph{A structural condition that pays off.}
The key structural assumption underpinning our decomposition and conditioning is that the three components of our noise model are mutually independent. Not only does this assumption translate to additivity in the relevant matrix identities, but, more importantly, it enables clean removal of pervasive effects when we condition on the pervasive component.

\paragraph{Contributions.}
\begin{itemize}[leftmargin=1.5em,itemsep=0.3em, parsep=0pt]
\item \textbf{D--C--L mixed-confounding formulation and precision decomposition.} We introduce a three-component Gaussian noise model composed of mutually independent (i) diagonal noise, (ii) localized low-rank confounders, and (iii) pervasive low-rank confounders. From that, we derive a D--C--L decomposition of the observed precision into a structured sparse component and a low-rank component.
\item \textbf{Structural characterization of the components.} We prove that the pervasive component is PSD and low-rank and that the non-pervasive component is exactly local under disjoint confounding support and approximately local under controlled overlap. We further characterize how this locality translates to the population-level precision induced by the model.
\item \textbf{Precision-led deconfounding pipeline (DCL--DECOR).} We propose a three-stage pipeline:
\begin{enumerate}
    \item estimate the structured--low-rank precision split via latent-variable graphical lasso with a configurable locality regularizer;
    \item invert the structured component to obtain a pervasive-adjusted (conditional) covariance estimate;
    \item run a correlated-noise continuous DAG learner (DECOR-GL) that jointly learns the DAG and the residual error precision, followed by a simple post-hoc bow reconciliation rule to enforce bow-freeness in the output.
\end{enumerate}
\item \textbf{Identifiability guarantees.} We show that, under standard transversality assumptions, the conditioned precision is identifiable from the population precision. Under additional mild conditions, we further identify the minimal bow-free target determined by the conditioned model; bow-freeness is a standard structural condition under which linear Gaussian SEM parameters become identifiable from observational covariances and has also been leveraged in recent bow-free covariance search procedures \citep{drton2011global, grassi2024sembap}.
\end{itemize}

\paragraph{Positioning with prior art.}
Constraint-based and score-based methods provide the classical backdrop \citep{spirtes2000causation,chickering2002optimal}. Continuous formulations such as NOTEARS and its descendants bring differentiability and scalability to DAG learning \citep{zheng2018dags}. Latent-variable graphical modeling separates sparse conditional structure from low-rank latent effects in undirected models \citep{chandrasekaran2012latent}. Our contribution is to \emph{bridge} these: use the precision domain to isolate pervasive directions and then invoke a DAG learner that explicitly models the remaining sparse confounding \citep{pal2025decor}. The result is a practical, provably grounded route to causal discovery in the linear Gaussian mixed-confounding regime—without interventions and without stepping outside the Gaussian world.

\section{Related Work}

\paragraph{Classical approaches without latent confounders.}
The modern literature on causal discovery for observational data begins with constraint-based procedures that leverage conditional independence (CI) tests to recover a Markov-equivalence class of DAGs. The PC algorithm and its variants are consistent under appropriate Markov and faithfulness assumptions when all relevant variables are observed \citep{spirtes2000causation}. Score-based search supplies a complementary view: Greedy Equivalence Search (GES) optimizes a penalized likelihood (e.g., BIC) over equivalence classes of DAGs and is consistent in large samples \citep{chickering2002optimal}. These families are powerful under \emph{causal sufficiency} but do not resolve directions within an equivalence class and typically degrade in the presence of unobserved confounding.

\paragraph{Latent variables and mixed graphs.}
When some causes are unobserved, the induced conditional independences on the observed margin can be represented by acyclic directed mixed graphs (ADMGs). Fast Causal Inference (FCI) and its relatives return a partial ancestral graph (PAG) encoding an equivalence class of ADMGs that is sound in the presence of hidden variables and selection bias \citep{spirtes2000causation,spirtes2001anytime,richardson2002ancestral}. For high-dimensional regimes and partial observability, RFCI improves sample efficiency by reducing the size of conditioning sets \citep{colombo2012learning}. Hybrid methods like GFCI \citep{ogarrio2016gfci} and PFCI \citep{pal2025penalized} combine a score step with CI testing and provide consistency guarantees while often improving empirical accuracy. Although these methods accommodate latent confounding, they generally leave many edges unoriented from a single observational environment in linear Gaussian settings.

\paragraph{Continuous optimization for DAG learning.}
A major development is to replace combinatorial search with continuous objectives that encode acyclicity via smooth constraints. NOTEARS uses a trace-exponential characterization to enforce DAGness and optimizes a penalized Gaussian least-squares score by gradient methods \citep{zheng2018dags}. Subsequent work broadened the modeling scope and improved scaling: nonparametric mechanisms and nonlinear SEMs \citep{zheng2020learning}, time-series structure via DYNOTEARS \citep{pamfil2020dynotears}, likelihood-based scores in GOLEM \citep{ng2020role}, and log-determinant-based acyclicity in DAGMA \citep{bello2022dagma}. While these methods scale gracefully and often achieve strong accuracy on curated benchmarks, they typically assume independent errors and can be sensitive to latent confounding.

\paragraph{Differentiable discovery under latent confounding.}
A line of work incorporates latent confounding directly into continuous formulations. \cite{bhattacharya2021differentiable} derive differentiable algebraic constraints for ADMGs and optimize a likelihood subject to mixed-graph structure, enabling discovery of bidirected edges. This treats latent dependence largely as \emph{dense} nuisance covariance (flexible but potentially under-identified) and does not exploit structured decompositions of the noise that enable identifiability in linear Gaussian models from a single environment. Other recent works in this direction include \citep{prashant2024differentiable,ma2024scalable}.

\paragraph{Low-rank\,+\,sparse decompositions for latent structure (undirected).}
In Gaussian graphical modeling, a now-standard approach separates conditional structure from shared latent effects by decomposing the \emph{precision} matrix as a sparse part minus a low-rank positive semidefinite part. The convex latent-variable graphical lasso of \citet{chandrasekaran2012latent} provides identifiability and consistency under incoherence and transversality conditions. Related matrix decompositions---most prominently robust PCA \citep{candes2011rpca}---formalize when a low-rank component can be separated from a sparse component. On the covariance side, factor-plus-sparse estimators such as POET \citep{fan2013poet} and the graphical lasso \citep{friedman2008glasso} support scalable estimation in high dimensions. These ideas, however, target undirected structure and do not, by themselves, orient edges.

\paragraph{Non-Gaussian and nonlinear identifiability.}
Stronger distributional assumptions can break the observational symmetry. In linear non-Gaussian SEMs, LiNGAM identifies a unique ordering and orientation from a single environment \citep{shimizu2006lingam}. Nonlinear additive-noise models can also be identifiable by exploiting asymmetries in functional mechanisms \citep{hoyer2009nonlinear}. With latent confounding, recent work shows that in linear \emph{non-Gaussian} models satisfying bow-free restrictions, the exact causal graph is identifiable \citep{wang2023nonGaussian}. These advances are compelling but move outside the linear Gaussian setting that underlies many pipelines in practice.

\paragraph{Position of the present work.}
Our setting is linear Gaussian with both \emph{pervasive} (dense low-rank) and \emph{sparse} low-rank confounding. We build on latent-variable precision decompositions from the undirected literature \citep{chandrasekaran2012latent,fan2013poet,friedman2008glasso,candes2011rpca} and marry them with differentiable DAG learning \citep{zheng2018dags,ng2020role,bello2022dagma,pamfil2020dynotears}. The key ingredient is to use the learned structured component of the precision to form a pervasive-adjusted covariance estimate, after which a DAG learner that explicitly models \emph{sparse} confounding can be applied \citep{pal2025decor}. This yields identifiability and a practical algorithm under homoskedastic idiosyncratic noise, incoherence of pervasive factors, bounded-degree sparsity, and a mild separated-touch (or dominance) condition—while remaining within the linear Gaussian world and a single observational environment.

\section{Problem Setup}
\label{sec:setup}

We consider a linear Gaussian structural equation model (SEM) on \(p\) observed variables \(\mathbf{x}\in\mathbb{R}^{p}\),
\begin{equation}
\label{eq:sem}
\mathbf{x} \;=\; \mathbf{B}^\top \mathbf{x} \;+\; \boldsymbol{\varepsilon},
\qquad 
\mathbf{T} \;:=\; \mathbf{I} - \mathbf{B},
\end{equation}
where \(\mathbf{B}\) encodes a directed acyclic graph (DAG) and \(\mathbf{T}\) is invertible under some causal ordering (unit-diagonal and triangular in that order), with \(\det(\mathbf{T})=1\). Writing \(\mathbf{x}=\mathbf{T}^{-\top}\boldsymbol{\varepsilon}\), all randomness is in the exogenous noise \(\boldsymbol{\varepsilon}\).
The population covariance and precision of \(\mathbf{x}\) are
\(
\boldsymbol{\Sigma} \;=\; \mathbf{T}^{-\top}\boldsymbol{\Omega}\,\mathbf{T}^{-1},
\,
\boldsymbol{\Theta} \;=\; \boldsymbol{\Sigma}^{-1} \;=\; \mathbf{T}\,\boldsymbol{\Omega}^{-1}\mathbf{T}^\top.
\)

\paragraph{Subscript convention (noise vs.\ observed level).}
We use a subscript \(\varepsilon\) for quantities defined at the \emph{noise level} (i.e., functions of \(\boldsymbol{\Omega}\) and \(\boldsymbol{\Omega}^{-1}\)) and a subscript \(x\) for their \emph{observed-level} counterparts (i.e., functions of \(\boldsymbol{\Sigma}\) and \(\boldsymbol{\Theta}\)). The two levels are related by congruence through \(\mathbf{T}\):
\(
\mathbf{M}_x \;:=\; \mathbf{T}\,\mathbf{M}_\varepsilon\,\mathbf{T}^\top
\, \text{for any noise-level matrix }\mathbf{M}_\varepsilon.
\)

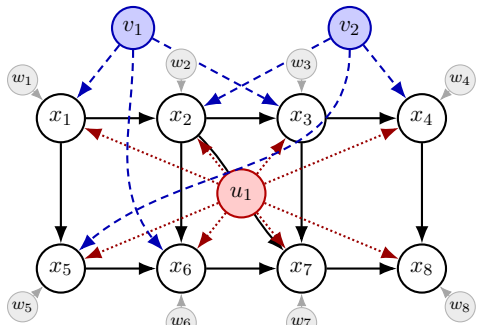
\begin{SCfigure}[1][t]
\begin{minipage}[t]{0.4\textwidth}
\begin{tikzpicture}[
  scale=0.8, transform shape,
  >=Latex,
  obs/.style={circle, draw=black, thick, fill=white, minimum size=8mm, inner sep=0pt},
  noise/.style={circle, draw=gray!60, fill=gray!15, minimum size=5mm, inner sep=0pt, font=\scriptsize},
  latS/.style={circle, draw=blue!70!black, thick, fill=blue!20, minimum size=7mm, inner sep=0pt},
  latL/.style={circle, draw=red!70!black, thick, fill=red!20, minimum size=8mm, inner sep=0pt},
  causal/.style={->, thick, black},
  confS/.style={->, densely dashed, blue!70!black, thick},
  confL/.style={->, densely dotted, red!60!black, thick},
  idio/.style={->, thin, gray!70}
]

\node[obs] (x1) at (0, 0)   {$x_1$};
\node[obs] (x2) at (2, 0)   {$x_2$};
\node[obs] (x3) at (4, 0)   {$x_3$};
\node[obs] (x4) at (6, 0)   {$x_4$};
\node[obs] (x5) at (0, -2.5) {$x_5$};
\node[obs] (x6) at (2, -2.5) {$x_6$};
\node[obs] (x7) at (4, -2.5) {$x_7$};
\node[obs] (x8) at (6, -2.5) {$x_8$};

\draw[causal] (x1) -- (x2);
\draw[causal] (x2) -- (x3);
\draw[causal] (x3) -- (x4);
\draw[causal] (x1) -- (x5);
\draw[causal] (x2) -- (x6);
\draw[causal] (x3) -- (x7);
\draw[causal] (x4) -- (x8);
\draw[causal] (x5) -- (x6);
\draw[causal] (x6) -- (x7);
\draw[causal] (x7) -- (x8);
\draw[causal] (x2) to[out=-45, in=135] (x7);

\node[latL] (u1) at (3, -1.25) {$u_1$};
\draw[confL] (u1) -- (x1);
\draw[confL] (u1) -- (x2);
\draw[confL] (u1) -- (x3);
\draw[confL] (u1) -- (x4);
\draw[confL] (u1) -- (x5);
\draw[confL] (u1) -- (x6);
\draw[confL] (u1) -- (x7);
\draw[confL] (u1) -- (x8);

\node[latS] (v1) at (1.2, 1.5) {$v_1$};
\draw[confS] (v1) -- (x1);
\draw[confS] (v1) -- (x3);
\draw[confS] (v1) to[out=-90, in=135] (x6);

\node[latS] (v2) at (4.8, 1.5) {$v_2$};
\draw[confS] (v2) -- (x2);
\draw[confS] (v2) -- (x4);
\draw[confS] (v2) to[out=-90, in=45] (x5);

\node[noise] (w1) at ($(x1)+(135:0.9)$) {$w_1$};
\node[noise] (w2) at ($(x2)+(90:0.9)$)  {$w_2$};
\node[noise] (w3) at ($(x3)+(90:0.9)$)  {$w_3$};
\node[noise] (w4) at ($(x4)+(45:0.9)$)  {$w_4$};
\node[noise] (w5) at ($(x5)+(-135:0.9)$) {$w_5$};
\node[noise] (w6) at ($(x6)+(-90:0.9)$)  {$w_6$};
\node[noise] (w7) at ($(x7)+(-90:0.9)$)  {$w_7$};
\node[noise] (w8) at ($(x8)+(-45:0.9)$)  {$w_8$};
\foreach \i in {1,...,8} {
  \draw[idio] (w\i) -- (x\i);
}
\end{tikzpicture}
\end{minipage}
\vspace{-24pt}
\caption{The D--C--L noise model. Observed variables $x_1,\ldots,x_8$ follow a DAG (solid arrows). 
Each receives idiosyncratic noise $w_i$ (gray). 
Localized confounders $v_j$ (blue, dashed) each affect three variables with no direct DAG edges among them (bow-free). 
The pervasive confounder $u_1$ (red, dotted) in the center affects all variables, inducing low-rank correction $\mathbf{L}_\varepsilon$.}
\label{fig:dcl-schematic}
\end{SCfigure}

\subsection{Noise Model and Mixed Confounding}
\label{subsec:noise}

In many real-world systems, unobserved confounders exhibit heterogeneous structure: some affect only small subsets of variables while others influence nearly all measurements. For instance, in gene expression, microRNAs can induce localized correlations across targeted gene sets, whereas chromatin state can act pervasively across the genome.

To capture this heterogeneity, we model the noise as a sum of three independent components,
\begin{equation}
\label{eq:noise}
\boldsymbol{\varepsilon} \;=\; 
\mathbf{W}\mathbf{w} \;+\; 
\mathbf{V}\mathbf{v} \;+\; 
\mathbf{U}\mathbf{u}, 
\quad 
\mathbf{w}\sim\mathcal{N}(\mathbf{0},\mathbf{I}_p),\ 
\mathbf{v}\sim\mathcal{N}(\mathbf{0},\mathbf{I}_{r_S}),\ 
\mathbf{u}\sim\mathcal{N}(\mathbf{0},\mathbf{I}_{r_L}),
\end{equation}
where \(\mathbf{W},\mathbf{V},\mathbf{U}\) encode how each type of noise acts on the observed variables. The diagonal matrix \(\mathbf{W}\in\mathbb{R}^{p\times p}\) represents heteroskedastic idiosyncratic noise at each node. The matrix \(\mathbf{V}\in\mathbb{R}^{p\times r_S}\) captures \emph{localized confounding}: each column \(\mathbf{V}_{\cdot j}\) has small support (e.g., \(|\mathrm{supp}(\mathbf{V}_{\cdot j})|\le s\ll p\)). Finally, \(\mathbf{U}\in\mathbb{R}^{p\times r_L}\) captures \emph{pervasive confounding} with dense loadings and \(r_L\ll p\).

The noise covariance is
\(
\boldsymbol{\Omega}
\;=\;
\mathrm{Var}(\boldsymbol{\varepsilon})
\;=\;
\mathbf{W}\mathbf{W}^\top
\;+\;
\mathbf{V}\mathbf{V}^\top
\;+\;
\mathbf{U}\mathbf{U}^\top.
\)

\subsection{Covariance, Precision, and a D--C--L Decomposition}
\label{subsec:Sigma-Theta}

Our first step is to analyze the structure of \(\boldsymbol{\Omega}^{-1}\). Define the diagonal idiosyncratic precision
\(
\mathbf{D}_\varepsilon
:=
(\mathbf{W}\mathbf{W}^\top)^{-1}
=
\mathrm{diag}(d_1,\ldots,d_p),\, d_i>0,
\)
and the overlap matrix
\(
\mathbf{A}
\;:=\;
\mathbf{I}+\mathbf{V}^\top\mathbf{D}_\varepsilon\mathbf{V}
\ \in\
\mathbb{R}^{r_S\times r_S}.
\)
Applying the Sherman--Morrison--Woodbury (SMW) identity to
\(\mathbf{W}\mathbf{W}^\top + \mathbf{V}\mathbf{V}^\top\) yields the \emph{structured non-pervasive precision}
\[
\mathbf{S}_\varepsilon
:=
(\mathbf{W}\mathbf{W}^\top+\mathbf{V}\mathbf{V}^\top)^{-1}
=
\mathbf{D}_\varepsilon
-
\underbrace{\mathbf{D}_\varepsilon\mathbf{V}\mathbf{A}^{-1}\mathbf{V}^\top\mathbf{D}_\varepsilon}_{\mathbf{C}_\varepsilon}.
\]
Here \(\mathbf{C}_\varepsilon\) (for \textbf{C}oupling) collects the pairwise dependencies induced by localized confounders. Importantly, the \emph{structured} object we aim to exploit is \(\mathbf{S}_\varepsilon=\mathbf{D}_\varepsilon-\mathbf{C}_\varepsilon\), which need not be strictly sparse: depending on the geometry of confounder supports, \(\mathbf{S}_\varepsilon\) can be row-sparse, banded, block-diagonal, etc.\ (see Remark~\ref{rem:beyond-sparsity} and Appendix~\ref{app:structure-pres}).

Now write \(\boldsymbol{\Omega}=\mathbf{S}_\varepsilon^{-1}+\mathbf{U}\mathbf{U}^\top\) and apply SMW again. Defining
\(
\mathbf{L}_\varepsilon
:=
\mathbf{S}_\varepsilon\mathbf{U}\big(\mathbf{I}+\mathbf{U}^\top\mathbf{S}_\varepsilon\mathbf{U}\big)^{-1}\mathbf{U}^\top\mathbf{S}_\varepsilon,
\)
the noise precision admits the D--C--L decomposition
\begin{equation}
\label{eq:OmegaInv}
\boldsymbol{\Omega}^{-1}
\;=\;
\underbrace{\mathbf{D}_\varepsilon}_{\text{diagonal}}
\;-\;
\underbrace{\mathbf{C}_\varepsilon}_{\text{localized coupling}}
\;-\;
\underbrace{\mathbf{L}_\varepsilon}_{\text{pervasive (low-rank)}}
\;=\;
\mathbf{S}_\varepsilon - \mathbf{L}_\varepsilon.
\end{equation}

This decomposition is not directly observable because we do not observe \(\boldsymbol{\Omega}\) itself. Crucially, however, the same split appears at the observed level.

\paragraph{Observed-level decomposition.}
Congruencing \eqref{eq:OmegaInv} by \(\mathbf{T}\) yields
\begin{equation}
\label{eq:Theta-DSL}
\boldsymbol{\Theta}
\;=\;
\mathbf{T}\boldsymbol{\Omega}^{-1}\mathbf{T}^\top
\;=\;
\underbrace{\mathbf{T}\mathbf{D}_\varepsilon\mathbf{T}^\top}_{\mathbf{D}_x}
\;-\;
\underbrace{\mathbf{T}\mathbf{C}_\varepsilon\mathbf{T}^\top}_{\mathbf{C}_x}
\;-\;
\underbrace{\mathbf{T}\mathbf{L}_\varepsilon\mathbf{T}^\top}_{\mathbf{L}_x}
\;=\;
\underbrace{\mathbf{T}\mathbf{S}_\varepsilon\mathbf{T}^\top}_{\mathbf{S}_x}
\;-\;
\mathbf{L}_x.
\end{equation}
The compact form \(\boldsymbol{\Theta}=\mathbf{S}_x-\mathbf{L}_x\) matches the standard \emph{structured-minus-low-rank} precision model used in latent graphical lasso approaches \citep{chandrasekaran2012latent}. This alignment motivates our estimation strategy: we estimate \((\mathbf{S}_x,\mathbf{L}_x)\) from data, then use \(\mathbf{S}_x\) (after deconfounding) to recover the DAG structure.

\begin{remark}[Beyond sparsity: other local structure classes]
\label{rem:beyond-sparsity}
While much of our algorithmic development focuses on row-sparse structure (amenable to \(\ell_1\)-regularization), the same decomposition supports other ``local'' structures for \(\mathbf{S}_\varepsilon\), such as banded or block-diagonal patterns induced by contiguous or group-confined confounder supports. The key requirement is that the structure class be (approximately) preserved under \(\mathbf{T}\)-congruence; see Proposition~\ref{prop:T-congruence} in Appendix~\ref{app:structure-pres}.
\end{remark}

\subsection{Relation to existing formulations}
\label{subsec:related-DSL}

The components in \eqref{eq:OmegaInv} recover several settings studied previously:
\begin{itemize}[leftmargin=1.2em, itemsep=0.3em, parsep=0pt]
\item \textbf{Independent errors (D only).} If \(\mathbf{V}=\mathbf{U}=\mathbf{0}\), then \(\mathbf{C}_\varepsilon=\mathbf{0}\), \(\mathbf{L}_\varepsilon=\mathbf{0}\), and
\(\boldsymbol{\Theta}=\mathbf{D}_x=\mathbf{T}\mathbf{D}_\varepsilon\mathbf{T}^\top\).
This reduces to a standard SEM with heteroskedastic but uncorrelated errors; \citet{loh2014high} relate the support of \(\mathbf{D}_x\) to the moralized graph.

\item \textbf{Pervasive confounding (D--L).} If \(\mathbf{V}=\mathbf{0}\), then \(\mathbf{C}_\varepsilon=\mathbf{0}\), \(\mathbf{S}_\varepsilon=\mathbf{D}_\varepsilon\), and \(\boldsymbol{\Theta}=\mathbf{D}_x-\mathbf{L}_x\), the sparse-minus-low-rank regime central to latent-variable Gaussian graphical models \citep{chandrasekaran2012latent} and confounded DAG learning methods such as \citet{frot2019robust,agrawal2023decamfounder}.

\item \textbf{Localized confounding (D--C).} If \(\mathbf{U}=\mathbf{0}\), then \(\mathbf{L}_\varepsilon=\mathbf{0}\) and \(\boldsymbol{\Theta}=\mathbf{S}_x=\mathbf{D}_x-\mathbf{C}_x\), corresponding to a DAG with localized (structured) error coupling. This setting is closely related to recent work such as DAG-DECOR \citep{pal2025decor}.

\item \textbf{Mixed confounding (D--C--L, this work).} In the full model, both localized coupling and pervasive latent factors co-occur, yielding \(\boldsymbol{\Theta}=\mathbf{S}_x-\mathbf{L}_x\) with \(\mathbf{S}_x=\mathbf{D}_x-\mathbf{C}_x\). Our goal is to develop theory and algorithms that handle this combined regime.
\end{itemize}

\subsection{Structural properties of precision matrix components}
\label{subsec:LS-structure-main}

We summarize the key properties that underpin our estimation procedure; full statements and proofs are given in Appendix~\ref{app:structure}.

\begin{proposition}[Low-rankness of \(\mathbf{L}_\varepsilon\) and locality of \(\mathbf{S}_\varepsilon\)]
\label{prop:lowrank-sparse-main}
Assume the model in \S\ref{sec:setup} with \(\mathbf{T}=\mathbf{I}-\mathbf{B}\) unit-diagonal and triangular under some causal order, and define \(\mathbf{D}_\varepsilon,\mathbf{C}_\varepsilon,\mathbf{S}_\varepsilon,\mathbf{L}_\varepsilon\) as above.
\begin{enumerate}[leftmargin=1.5em, itemsep=0.3em, parsep=0pt]
\item[\textnormal{(a)}] \textbf{Low-rank, PSD pervasive correction.}  
\(\mathbf{L}_\varepsilon\succeq 0\) and \(\mathrm{rank}(\mathbf{L}_\varepsilon)\le r_L\). Consequently, \(\mathbf{L}_x=\mathbf{T}\mathbf{L}_\varepsilon\mathbf{T}^\top\succeq 0\) with \(\mathrm{rank}(\mathbf{L}_x)\le r_L\).
\item[\textnormal{(b)}] \textbf{Exact locality under disjoint supports.}  
If the columns of \(\mathbf{V}\) have disjoint supports of size at most \(s\), and each variable participates in at most \(c\) such supports, then each row of \(\mathbf{S}_\varepsilon\) has at most \(c(s-1)\) off-diagonal nonzeros (equivalently, \(\mathbf{C}_\varepsilon\) is supported on a union of small cliques).
\item[\textnormal{(c)}] \textbf{Controlled leakage under overlap.}  
If the confounder supports overlap mildly, then \(\mathbf{S}_\varepsilon\) remains \emph{approximately local}: each row has only \(\mathcal{O}(cs)\) entries above a fixed threshold, provided the overlap-induced leakage is controlled. A sufficient condition is given by Proposition~\ref{prop:leakage} (Appendix~\ref{app:leakage}).
\item[\textnormal{(d)}] \textbf{Propagation through the DAG.}  
If the DAG \(\mathbf{B}\) has bounded degree, then \(\mathbf{S}_x=\mathbf{T}\mathbf{S}_\varepsilon\mathbf{T}^\top\) preserves locality up to a constant inflation factor (row-sparsity in the sparse case; bandedness/block-structure in ordered or grouped settings as in Proposition~\ref{prop:T-congruence}).
\end{enumerate}
\end{proposition}

\section{From Precision Decomposition to a Deconfounding Algorithm}
\label{sec:pop-identities}

At the population level, \S\ref{sec:setup} implies an observed-level precision decomposition
\begin{equation}
\label{eq:Theta-SxLx}
\boldsymbol{\Theta}
=
\boldsymbol{\Sigma}^{-1}
=
\underbrace{\mathbf{T}\mathbf{S}_\varepsilon\mathbf{T}^\top}_{\mathbf{S}_x}
\;-\;
\underbrace{\mathbf{T}\mathbf{L}_\varepsilon\mathbf{T}^\top}_{\mathbf{L}_x},
\qquad
\mathbf{S}_\varepsilon
=
(\mathbf{W}\mathbf{W}^\top+\mathbf{V}\mathbf{V}^\top)^{-1}
=
\boldsymbol{\Gamma}_\varepsilon^{-1},
\end{equation}
where $\mathbf{L}_x\succeq 0$ has rank at most $r_L$, while $\mathbf{S}_x$ inherits a \emph{local} structure from $\mathbf{S}_\varepsilon$ under bounded-degree $\mathbf{T}$-congruence (Proposition~\ref{prop:lowrank-sparse-main}).
The decomposition~\eqref{eq:Theta-SxLx} has a direct probabilistic interpretation: $\mathbf{S}_x$ is the precision matrix of $\mathbf{x}$ after conditioning on the pervasive factors $\mathbf{u}$.

\begin{proposition}[Conditional precision]
\label{prop:conditional-precision}
Under \eqref{eq:sem}--\eqref{eq:noise}, let $\boldsymbol{\Gamma}_\varepsilon:=\mathbf{W}\mathbf{W}^\top+\mathbf{V}\mathbf{V}^\top$ and $\mathbf{S}_\varepsilon=\boldsymbol{\Gamma}_\varepsilon^{-1}$. Then
$
\mathrm{Cov}(\mathbf{x}\mid \mathbf{u})
=
\mathbf{T}^{-\top}\boldsymbol{\Gamma}_\varepsilon\mathbf{T}^{-1},
\,
\mathrm{Cov}(\mathbf{x}\mid \mathbf{u})^{-1}
=
\mathbf{T}\mathbf{S}_\varepsilon\mathbf{T}^\top
=
\mathbf{S}_x.$

\end{proposition}
Proposition~\ref{prop:conditional-precision} reduces mixed confounding to a correlated-noise SEM: after conditioning on $\mathbf{u}$, the DAG $\mathbf{B}$ is unchanged and the remaining noise covariance $\boldsymbol{\Gamma}_\varepsilon$ encodes only idiosyncratic and localized confounding.

\subsection{A three-stage pipeline}
\label{subsec:pipeline}

We operationalize the reduction in three stages: (I)~estimate the structured--low-rank precision split, (II)~invert the structured component to obtain the deconfounded covariance, and (III)~learn the DAG via a correlated-noise estimator.

\paragraph{Stage I: Structured--low-rank precision split.}
Given the sample covariance $\widehat{\boldsymbol{\Sigma}} = n^{-1}\mathbf{X}^\top\mathbf{X}$, we estimate $(\mathbf{S}_x,\mathbf{L}_x)$ via the latent-variable graphical lasso \citep{chandrasekaran2012latent}:
\begin{equation}
\label{eq:lvglasso}
(\widehat{\mathbf{S}}_x,\widehat{\mathbf{L}}_x)
\;\in\;
\argmin_{\substack{\mathbf{S}\succ 0,\,\mathbf{L}\succeq 0\\ \mathbf{S}-\mathbf{L}\succ 0}}
\Big\{
-\log\det(\mathbf{S}-\mathbf{L})
+\mathrm{tr}\!\big(\widehat{\boldsymbol{\Sigma}}(\mathbf{S}-\mathbf{L})\big)
+\lambda_s\,\mathcal{R}_{\mathrm{loc}}(\mathbf{S})
+\lambda_*\mathrm{tr}(\mathbf{L})
\Big\},
\end{equation}
where $\mathcal{R}_{\mathrm{loc}}$ is a convex regularizer encoding the chosen locality class (e.g., $\|\mathbf{S}\|_{1,\mathrm{off}}$ for sparsity; see Remark~\ref{rem:beyond-sparsity} for banded and block-sparse alternatives).

\paragraph{Stage II: Deconfounded covariance.}
By Proposition~\ref{prop:conditional-precision}, the conditional covariance $\boldsymbol{\Sigma}_{\mathrm{cond}} = \mathrm{Cov}(\mathbf{x} \mid \mathbf{u})$ equals $\mathbf{S}_x^{-1}$. We estimate it by inverting the structured component:
$\widehat{\boldsymbol{\Sigma}}_{\mathrm{cond}}
\;:=\;
\widehat{\mathbf{S}}_x^{-1}.
$
This inversion is computed via sparse Cholesky factorization and triangular solves, exploiting the local structure of $\widehat{\mathbf{S}}_x$ without forming an explicit dense inverse.

\paragraph{Stage III: Correlated-noise DAG learning.}
By Proposition~\ref{prop:conditional-precision}, conditioning on $\mathbf{u}$ removes pervasive confounding at the second-moment level. The \emph{conditional residual}
$
\mathbf{x}^\perp := \mathbf{x}-\mathbb{E}[\mathbf{x}\mid \mathbf{u}]
$
obeys the same linear SEM with \emph{correlated} errors driven only by idiosyncratic and localized confounding:
\begin{equation}
\label{eq:cond-sem}
\mathbf{x}^\perp
=
\mathbf{B}^\top\mathbf{x}^\perp + \boldsymbol{\varepsilon}^{\perp},
\qquad
\boldsymbol{\varepsilon}^{\perp}\sim\mathcal{N}(\mathbf{0},\boldsymbol{\Gamma}_\varepsilon),
\qquad
\boldsymbol{\Sigma}_{\mathrm{cond}}
=
\mathbf{T}^{-\top}\boldsymbol{\Gamma}_\varepsilon\mathbf{T}^{-1},
\end{equation}
where $\mathbf{T}=\mathbf{I}-\mathbf{B}$ and $\mathbf{S}_\varepsilon=\boldsymbol{\Gamma}_\varepsilon^{-1}$.

We instantiate Stage~III with DECOR-GL \citep{pal2025decor}, a precision-penalized variant that directly enforces sparsity in $\mathbf{S}_\varepsilon$:
\begin{equation}
\label{eq:decor-gl-covform}
\min_{\mathbf{B},\,\mathbf{S}_\varepsilon\succ 0}\;
\underbrace{
\mathrm{tr}\!\big(\widehat{\boldsymbol{\Sigma}}_{\mathrm{cond}}\,\mathbf{T}\mathbf{S}_\varepsilon\mathbf{T}^\top\big)
-\log\det\mathbf{S}_\varepsilon
}_{\text{negative log-likelihood}}
\;+\;
\lambda_B\|\mathbf{B}\|_1
\;+\;
\lambda_S\|\mathbf{S}_\varepsilon\|_{1,\mathrm{off}}
\;+\;
\rho\,h(\mathbf{B}),
\end{equation}
where $h(\mathbf{B})=\mathrm{tr}(e^{\mathbf{B}\odot\mathbf{B}})-p$ is the smooth acyclicity surrogate \citep{zheng2018dags}. For any acyclic $\mathbf{B}$, $\mathbf{T}$ is unit-triangular with $\det(\mathbf{T})=1$, so no log-determinant term in $\mathbf{T}$ appears.

\begin{remark}[Connection to NOTEARS]
\label{rem:notears-connection}
Interpreting $\widehat{\boldsymbol{\Sigma}}_{\mathrm{cond}}$ as the empirical covariance of deconfounded samples $\mathbf{X}^\perp \in \mathbb{R}^{n \times p}$, the trace term satisfies
\(
\mathrm{tr}\!\big(\widehat{\boldsymbol{\Sigma}}_{\mathrm{cond}}\,(\mathbf{I}-\mathbf{B})\mathbf{S}_\varepsilon(\mathbf{I}-\mathbf{B})^\top\big)
\;\approx\;
\frac{1}{n}\big\|(\mathbf{X}^\perp-\mathbf{X}^\perp\mathbf{B})\,\mathbf{S}_\varepsilon^{1/2}\big\|_F^2.
\)
When $\mathbf{S}_\varepsilon = \mathbf{I}$ (independent errors), the $\mathbf{B}$-update reduces to NOTEARS. Stage~III thus generalizes NOTEARS to correlated-error SEMs, where the residual whitening matrix is learned jointly.
\end{remark}

\paragraph{DECOR-GL alternation.}
Problem~\eqref{eq:decor-gl-covform} is solved by alternating updates:
\begin{itemize}[leftmargin=1.2em,itemsep=0.3em]
\item \textbf{Graph step (given $\mathbf{S}_\varepsilon$).}
With $\mathbf{S}_\varepsilon$ fixed, update $\mathbf{B}$ by proximal-gradient descent on
\begin{equation}
\label{eq:decor-gl-step1}
\min_{\mathbf{B}}\;
\mathrm{tr}\!\big(\widehat{\boldsymbol{\Sigma}}_{\mathrm{cond}}\,(\mathbf{I}-\mathbf{B})\mathbf{S}_\varepsilon(\mathbf{I}-\mathbf{B})^\top\big)
\;+\;
\lambda_B\|\mathbf{B}\|_1
\;+\;
\rho\,h(\mathbf{B}),
\end{equation}
using an augmented-Lagrangian schedule for $h(\mathbf{B}) \approx 0$ as in NOTEARS.

\item \textbf{Noise step (given $\mathbf{B}$).}
With $\mathbf{B}$ fixed, form the residual covariance
$
\widehat{\boldsymbol{\Sigma}}_{\mathbf{B}}
:=
\mathbf{T}^\top \widehat{\boldsymbol{\Sigma}}_{\mathrm{cond}}\, \mathbf{T}
$
and solve the graphical lasso
\begin{equation}
\label{eq:glasso-stage3}
\widehat{\mathbf{S}}_\varepsilon
\;\in\;
\argmin_{\mathbf{S}\succ 0}
\Big\{
-\log\det\mathbf{S}
+\mathrm{tr}(\widehat{\boldsymbol{\Sigma}}_{\mathbf{B}}\,\mathbf{S})
+\lambda_S\|\mathbf{S}\|_{1,\mathrm{off}}
\Big\}.
\end{equation}
\end{itemize}

\paragraph{Post-hoc bow reconciliation.}
A \emph{bow} occurs when a pair $(i,j)$ has both a directed edge ($B_{ij} \neq 0$ or $B_{ji} \neq 0$) and a bidirected edge ($[\boldsymbol{\Gamma}_\varepsilon]_{ij} \neq 0$). Bows create fundamental non-identifiability even in two-node models (see \S\ref{subsec:ident-stage3}). Following \citet{pal2025decor}, we apply a post-hoc reconciliation step after convergence at iteration $t$. For each pair $(i,j)$ with both edge types present after thresholding, we compare the directed signal strength against the normalized error correlation:
\begin{equation}
\label{eq:bow-reconcile}
\text{keep directed edge} \;\Longleftrightarrow\;
\max\{|B^{(t)}_{ij}|, |B^{(t)}_{ji}|\}
\;\ge\;
c \cdot {|\Gamma^{(t)}_{\varepsilon,ij}|}/{\sqrt{\Gamma^{(t)}_{\varepsilon,ii}\,\Gamma^{(t)}_{\varepsilon,jj}}},
\end{equation}
where $c > 0$ is a tuning constant (we use $c = 1$). The weaker channel is zeroed, ensuring bow-freeness in the final output without complicating the optimization.

\begin{algorithm2e}[t]
\DontPrintSemicolon
\SetAlgoLined
\caption{\textsc{DCL-DECOR}: Deconfounding via D--C--L Decomposition}
\label{alg:dcl-decor}
\footnotesize 
\KwIn{Data $\mathbf{X}\in\mathbb{R}^{n\times p}$; penalties $\lambda_s, \lambda_*, \lambda_B, \lambda_S$; thresholds $\tau_B, \tau_\Gamma$; reconciliation constant $c > 0$.}
\BlankLine
\textbf{Stage I: Structured--low-rank precision split}\;
\Indp
Compute $\widehat{\boldsymbol{\Sigma}} \leftarrow n^{-1}\mathbf{X}^\top\mathbf{X}$\;
Solve LVGLASSO~\eqref{eq:lvglasso} for $(\widehat{\mathbf{S}}_x, \widehat{\mathbf{L}}_x)$\;
\Indm
\BlankLine
\textbf{Stage II: Deconfounded covariance}\;
\Indp
Compute $\widehat{\boldsymbol{\Sigma}}_{\mathrm{cond}} \leftarrow \widehat{\mathbf{S}}_x^{-1}$ via Cholesky factorization\;
\Indm
\BlankLine
\textbf{Stage III: DECOR-GL}\;
\Indp
Initialize $\mathbf{S}^{(0)}_\varepsilon \leftarrow \mathrm{diag}(\widehat{\mathbf{S}}_x)$, $\mathbf{B}^{(0)} \leftarrow \mathbf{0}$\;
\Repeat{convergence at iteration $t$}{
  \textbf{Graph step:} Update $\mathbf{B}^{(k+1)}$ via~\eqref{eq:decor-gl-step1} (proximal augmented Lagrangian)\;
  \textbf{Noise step:} Form $\widehat{\boldsymbol{\Sigma}}_{\mathbf{B}} \leftarrow (\mathbf{T}^{(k+1)})^\top\widehat{\boldsymbol{\Sigma}}_{\mathrm{cond}}\,\mathbf{T}^{(k+1)}$; solve~\eqref{eq:glasso-stage3} for $\mathbf{S}^{(k+1)}_\varepsilon$\;
}
\textbf{Post-processing:}\;
Hard-threshold: $\widetilde{\mathbf{B}} \leftarrow \mathsf{HT}(\mathbf{B}^{(t)}; \tau_B)$, $\widetilde{\boldsymbol{\Gamma}}_\varepsilon \leftarrow \mathsf{HT}_{\mathrm{off}}((\mathbf{S}^{(t)}_\varepsilon)^{-1}; \tau_\Gamma)$\;
Bow reconciliation: $(\widehat{\mathbf{B}}, \widehat{\boldsymbol{\Gamma}}_\varepsilon) \leftarrow$ apply~\eqref{eq:bow-reconcile} to $(\widetilde{\mathbf{B}}, \widetilde{\boldsymbol{\Gamma}}_\varepsilon)$ for all conflicting pairs\;
\Indm
\KwOut{DAG estimate $\widehat{\mathbf{B}}$; structured noise covariance $\widehat{\boldsymbol{\Gamma}}_\varepsilon$.}
\end{algorithm2e}

\section{Identifiability and Modular Guarantees}
\label{sec:identifiability}

This section establishes what is identifiable from the observed covariance $\boldsymbol{\Sigma}$ in the D--C--L model and characterizes the recovery target for Stage~III once pervasive confounding has been removed.

\subsection{Identifiability of the structured--low-rank split}
\label{subsec:ident-split}

The observed precision satisfies $\boldsymbol{\Theta} = \mathbf{S}_x - \mathbf{L}_x$ with $\mathbf{L}_x \succeq 0$ low-rank and $\mathbf{S}_x$ in a local structure class (sparse, banded, block-sparse, etc.). Under standard transversality conditions between the structured and low-rank tangent cones \citep{chandrasekaran2012latent}, the pair $(\mathbf{S}_x, \mathbf{L}_x)$ is identifiable from $\boldsymbol{\Theta}$.

\begin{assumption}[Transversality]
\label{ass:transversality}
The tangent cones of the structured class $\mathcal{S}$ and the low-rank manifold $\mathcal{L}_{r_L}$ at the true parameters intersect trivially:
$
\mathrm{T}_{\mathcal{S}}(\mathbf{S}_x) \cap \mathrm{T}_{\mathcal{L}_{r_L}}(\mathbf{L}_x) = \{\mathbf{0}\}.
$
\end{assumption}

Under Assumption~\ref{ass:transversality} LVGLASSO consistently recovers $(\mathbf{S}_x, \mathbf{L}_x)$, yielding estimation error $\delta_{S,n} := \|\widehat{\mathbf{S}}_x - \mathbf{S}_x\|_2 \to 0$ as $n \to \infty$ \citep{chandrasekaran2012latent}.

\subsection{Stability of the conditional covariance}
\label{subsec:ident-inversion}

The inversion $\boldsymbol{\Sigma}_{\mathrm{cond}} = \mathbf{S}_x^{-1}$ is Lipschitz-stable on well-conditioned SPD matrices:

\begin{lemma}[Inversion perturbation bound]
\label{lem:inv-perturb}
If $\|\widehat{\mathbf{S}}_x - \mathbf{S}_x\|_2 < \lambda_{\min}(\mathbf{S}_x)$, then
$
\big\|
\widehat{\boldsymbol{\Sigma}}_{\mathrm{cond}} - \boldsymbol{\Sigma}_{\mathrm{cond}}
\big\|_2
\;\le\;
\frac{\|\mathbf{S}_x^{-1}\|_2^2 \cdot \|\widehat{\mathbf{S}}_x - \mathbf{S}_x\|_2}{1 - \|\mathbf{S}_x^{-1}\|_2 \cdot \|\widehat{\mathbf{S}}_x - \mathbf{S}_x\|_2}.$
\end{lemma}
\noindent
The proof follows from standard matrix perturbation theory (Appendix~\ref{app:proof-inv-perturb}). In words: if Stage~I achieves small error $\delta_{S,n}$ and $\mathbf{S}_x$ is well-conditioned (bounded condition number $\kappa(\mathbf{S}_x) = \lambda_{\max}(\mathbf{S}_x)/\lambda_{\min}(\mathbf{S}_x)$), then $\widehat{\boldsymbol{\Sigma}}_{\mathrm{cond}}$ is a stable estimate of $\boldsymbol{\Sigma}_{\mathrm{cond}}$.

\subsection{Identifiability from the conditional covariance}
\label{subsec:ident-stage3}

After conditioning on $\mathbf{u}$, the model~\eqref{eq:cond-sem} is a linear Gaussian SEM with correlated errors. Such models are represented as \emph{acyclic directed mixed graphs} (ADMGs) $G = (V, E^{\to}, E^{\leftrightarrow})$, where $E^{\to} = \mathrm{supp}(\mathbf{B})$ encodes directed edges and $E^{\leftrightarrow} = \mathrm{supp}_{\mathrm{off}}(\boldsymbol{\Gamma}_\varepsilon)$ encodes bidirected edges from error correlation.

\paragraph{Bow obstruction.}
A \emph{bow} on pair $(i,j)$ occurs when both a directed edge ($B_{ij} \neq 0$ or $B_{ji} \neq 0$) and a bidirected edge ($[\boldsymbol{\Gamma}_\varepsilon]_{ij} \neq 0$) are present. Bows create fundamental non-identifiability that motivates the bow-free condition at the conditional level:
$\mathrm{supp}(\mathbf{B}) \;\cap\; \mathrm{supp}_{\mathrm{off}}(\boldsymbol{\Gamma}_\varepsilon) = \varnothing.$


\paragraph{Connection to the D--C--L model.}
In terms of the loading matrix $\mathbf{V}$, bow-freeness requires that no localized confounder simultaneously affects a parent-child pair in the DAG. Formally, for each column $\mathbf{v}_k$ of $\mathbf{V}$:
\(
\text{if } v_{ik} \neq 0 \text{ and } v_{jk} \neq 0, \text{ then } B_{ij} = B_{ji} = 0
\) (Figure~\ref{fig:dcl-schematic}.)

\paragraph{Identifiability within a fixed bow-free ADMG.}
For a fixed bow-free acyclic mixed graph $G$, \citet{drton2011global} established that the covariance map $(\mathbf{B}, \boldsymbol{\Gamma}_\varepsilon) \mapsto \boldsymbol{\Sigma}_{\mathrm{cond}}$ is injective on the parameter space restricted to $G$'s zero pattern, provided a uniform eigenvalue margin $\boldsymbol{\Gamma}_\varepsilon \succeq m\mathbf{I}$ for some $m > 0$. This supplies the identifiability foundation for Stage~III.

\paragraph{Identifiability across graphs: bow-free equivalence classes.}
Even when parameters are identifiable \emph{given} a graph, the same $\boldsymbol{\Sigma}_{\mathrm{cond}}$ can arise from multiple distinct bow-free ADMGs (distributional equivalence). Let
$
\mathcal{E}_{\mathrm{bow}}(\boldsymbol{\Sigma}_{\mathrm{cond}})
:=
\big\{(\mathbf{B}, \boldsymbol{\Gamma}_\varepsilon) :
\mathbf{B} \text{ acyclic},\;
\boldsymbol{\Gamma}_\varepsilon \succ 0,\;
(\mathbf{B}, \boldsymbol{\Gamma}_\varepsilon) \text{ bow-free},\;
\boldsymbol{\Sigma}(\mathbf{B}, \boldsymbol{\Gamma}_\varepsilon) = \boldsymbol{\Sigma}_{\mathrm{cond}}
\big\}.$
Following \citet{pal2025decor}, the natural estimation target is a \emph{minimal} (sparsest) representative:
$
\mathcal{E}_{\mathrm{bow}}^{\min}(\boldsymbol{\Sigma}_{\mathrm{cond}})
:=
\argmin_{(\mathbf{B}, \boldsymbol{\Gamma}_\varepsilon) \in \mathcal{E}_{\mathrm{bow}}(\boldsymbol{\Sigma}_{\mathrm{cond}})}
\big(
\|\mathbf{B}\|_0 + \|[\boldsymbol{\Gamma}_\varepsilon]_{\mathrm{off}}\|_0
\big).$
When $\boldsymbol{\Gamma}_\varepsilon$ is diagonal, $\mathcal{E}_{\mathrm{bow}}$ reduces to the standard Markov equivalence class of DAGs.

\paragraph{What DECOR-GL estimates.}
Stage~III (DECOR-GL) is a sparsity-regularized likelihood method over $(\mathbf{B}, \mathbf{S}_\varepsilon)$. Because $\mathcal{E}_{\mathrm{bow}}(\boldsymbol{\Sigma}_{\mathrm{cond}})$ may contain multiple elements, the statistically meaningful target is $\mathcal{E}_{\mathrm{bow}}^{\min}(\boldsymbol{\Sigma}_{\mathrm{cond}})$ rather than a unique ground-truth DAG. Under suitable initialization and regularization, DECOR-GL with post-hoc bow reconciliation consistently finds an element of $\mathcal{E}_{\mathrm{bow}}^{\min}$ \citep{pal2025decor}.

\subsection{End-to-end modular guarantee}
\label{subsec:reduction}

The D--C--L model admits a clean modular target: first recover the pervasive-adjusted (conditional) covariance
$\boldsymbol{\Sigma}_{\mathrm{cond}}=\mathbf{S}_x^{-1}$ from the structured--low-rank precision split, then recover the appropriate bow-free correlated-noise SEM object from $\boldsymbol{\Sigma}_{\mathrm{cond}}$.

\begin{theorem}[Modular reduction]
\label{thm:reduction}
Assume \emph{(split identifiability)} $(\mathbf{S}_x,\mathbf{L}_x)$ is identifiable from $\boldsymbol{\Sigma}$ under Assumption~\ref{ass:transversality};
\emph{(conditioning)} $\mathbf{S}_x$ is well-conditioned ($\kappa(\mathbf{S}_x)<\infty$);
and \emph{(bow-freeness)} the conditional SEM~\eqref{eq:cond-sem} satisfies bow-freeness. 
Then the identifiable causal target of the D--C--L model from $\boldsymbol{\Sigma}$ is the minimal bow-free equivalence class
$\mathcal{E}_{\mathrm{bow}}^{\min}(\boldsymbol{\Sigma}_{\mathrm{cond}})$. The proof is given in Appendix~\ref{app:proof-reduction}.
\end{theorem}

\begin{theorem}[End-to-end consistency, informal]
\label{thm:end2end}
Under the assumptions of Theorem~\ref{thm:reduction}, suppose Stage~I is consistent in operator norm,
$\delta_{S,n}=\|\widehat{\mathbf{S}}_x-\mathbf{S}_x\|_2\to 0$ as $n\to\infty$, and the Stage~III procedure is stable (continuous) to small perturbations of its covariance input around $\boldsymbol{\Sigma}_{\mathrm{cond}}$.
Then $\widehat{\boldsymbol{\Sigma}}_{\mathrm{cond}}=\widehat{\mathbf{S}}_x^{-1}\to \boldsymbol{\Sigma}_{\mathrm{cond}}$, and the Stage~III output converges to an element of
$\mathcal{E}_{\mathrm{bow}}^{\min}(\boldsymbol{\Sigma}_{\mathrm{cond}})$. The proof is provided in Appendix~\ref{app:proof-end2end}.
\end{theorem}

\section{Synthetic Experiments: Mixed Confounding (Pervasive Rank/Strength)}
\label{sec:exp-dcl1}

We evaluate causal discovery under \emph{mixed confounding}, where a few \emph{pervasive} latent factors affect many variables while additional \emph{localized} latent factors induce structured, sparse dependence.

\paragraph{Baselines.}
We compare our three-stage \textsc{DCL--DECOR-GL} pipeline against: (i)~\textsc{DECOR-GL}, which learns a DAG with correlated noise under bow-free identifiability \citep{pal2025decor}; (ii)~\textsc{DECAMF-LIN}, a factor/residualization-style baseline for \emph{pervasive} confounding (provided the true latent rank in our sweep) \citep{agrawal2023decamfounder}; (iii)~continuous, causal-sufficiency DAG learners \textsc{NOTEARS} and \textsc{GOLEM} \citep{zheng2018dags,ng2020role}; and (iv)~classical score/ICA baselines \textsc{GES} and \textsc{LiNGAM} \citep{chickering2002optimal,shimizu2006lingam}. We report directed-edge recovery under a shared thresholding rule and emphasize robustness as pervasive and localized confounding vary.

\paragraph{Simulator.}
We simulate linear Gaussian SEMs on $p=40$ observed variables,
$
\mathbf{x}=\mathbf{B}^\top\mathbf{x}+\boldsymbol{\varepsilon},
$
where $\mathbf{B}$ is acyclic with approximately $8\%$ directed edge density and edge weights sampled with random sign and magnitude $\mathrm{Unif}(0.5,2)$.
The exogenous noise follows the D--C--L mixed-confounding model
$
\boldsymbol{\varepsilon}=\mathbf{V}\mathbf{v}+\mathbf{U}\mathbf{u}+\mathbf{w},
$
where $\mathbf{w}\sim\mathcal{N}(0,0.36\,\mathbf{I})$, localized confounding is represented by $\mathbf{V}\in\mathbb{R}^{p\times q}$ with $q=15$ column-sparse loadings (each column has 6 active entries drawn from $\mathcal{N}(0,0.3^2)$), and pervasive confounding is represented by $\mathbf{U}\in\mathbb{R}^{p\times q_P}$ with dense loadings sampled as
$
U_{ik}\sim \mathcal{N}\!\left(0,(U_d/\sqrt{p})^2\right).
$
This scaling makes each pervasive factor have $\ell_2$-norm on the order of $U_d$, so the overall magnitude of the pervasive covariance contribution grows roughly with $q_P U_d^2$.
To align with our identifiability target, we enforce bow-freeness \emph{only with respect to localized confounders}: the simulator removes bows induced by $\mathbf{V}$, but does not remove bows induced by the pervasive component $\mathbf{U}$.

\paragraph{DCL1: sweeping pervasive rank and strength.}
We vary pervasive rank and strength over the grid
$
q_P\in\{1,3,5\},\;
U_d\in\{0.5,1.0,2.0\},
$
using $n=600$ samples and 10 replicates per grid cell, running all methods listed above.
All methods use standardized inputs and a shared reporting threshold $\tau=0.30$ to form the estimated directed support.
For fairness, all non-DCL methods use $\lambda_B=0.10$; \textsc{DCL--DECOR-GL} uses a DCL-specific $\lambda_B=0.01$ (as in our synthetic setting), while sharing $\lambda_\Theta=0.01$ with \textsc{DECOR-GL}.
\textsc{DCL--DECOR-GL} first estimates a structured--low-rank precision split (LVGLASSO) with $(\lambda_{S_x},\lambda_{L_x})=(0.001,0.005)$, then inverts the structured component to obtain a pervasive-adjusted covariance estimate and runs \textsc{DECOR-GL} on that input.
Both \textsc{DECOR-GL} and \textsc{DCL--DECOR-GL} use identical post-hoc bow reconciliation (strict mode with the same thresholds and constant).

\begin{figure}[t]
  \centering
  \textbf{(a) DCL1: }$\Delta$F1 vs.\ DECOR-GL (mean over 10 reps)\par
  \includegraphics[width=.8\textwidth]{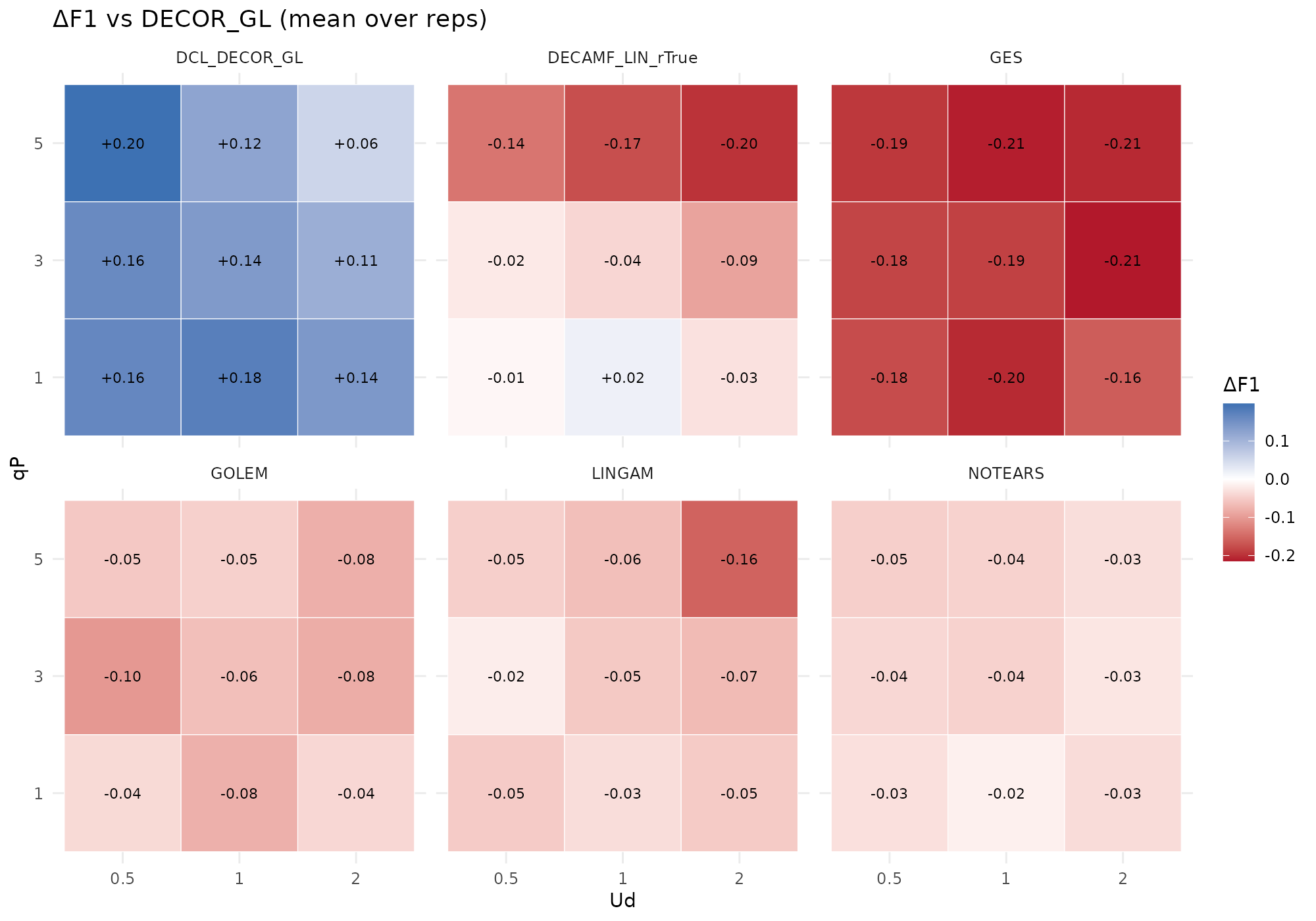}\vspace{0.5em}

  \begin{minipage}{0.49\textwidth}
    \centering
    \textbf{(b) DCL2: F1 vs.\ localized density}\par
    \includegraphics[width=\textwidth]{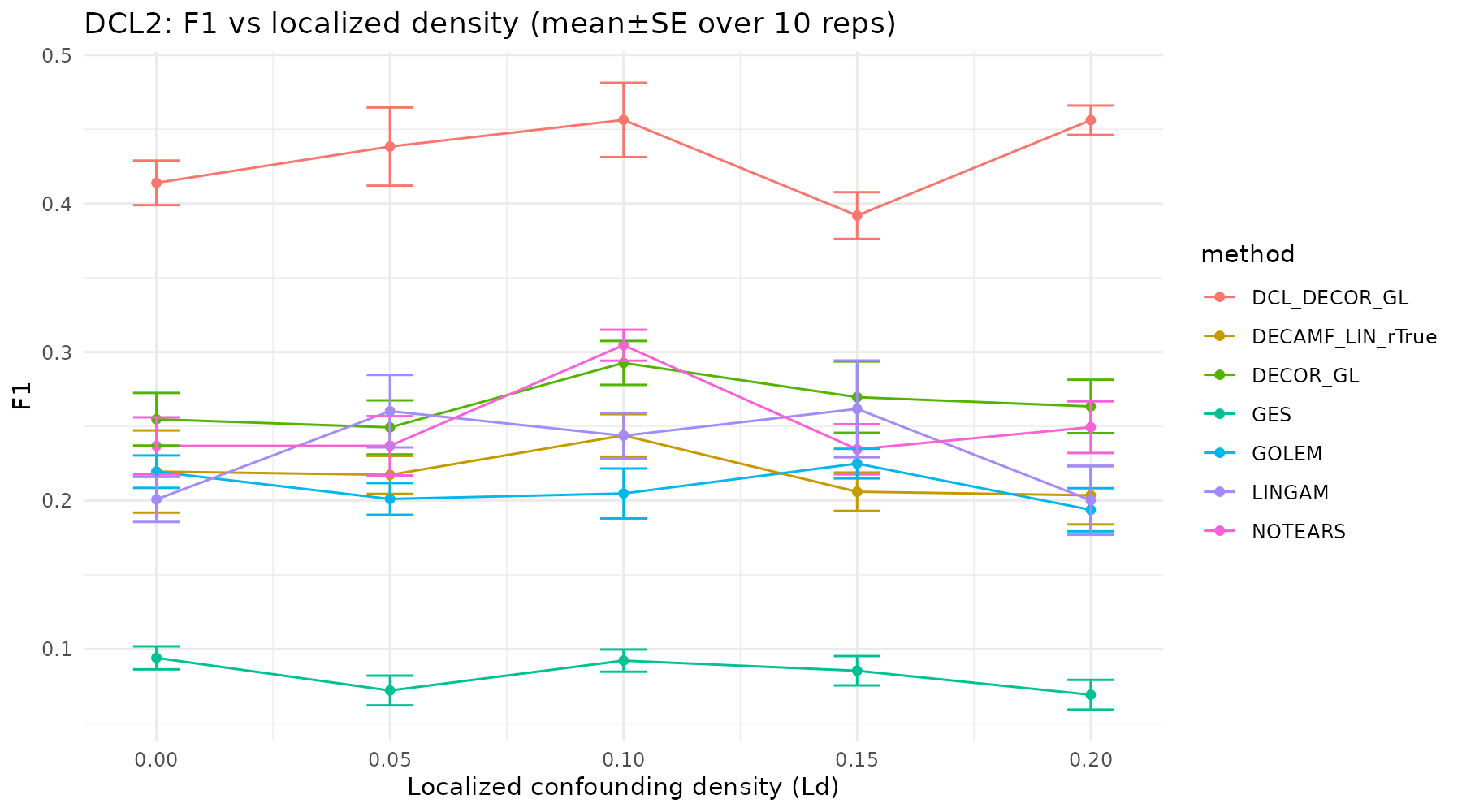}
  \end{minipage}\hfill
  \begin{minipage}{0.49\textwidth}
    \centering
    \textbf{(c) DCL2: SHD vs.\ localized density}\par
    \includegraphics[width=\textwidth]{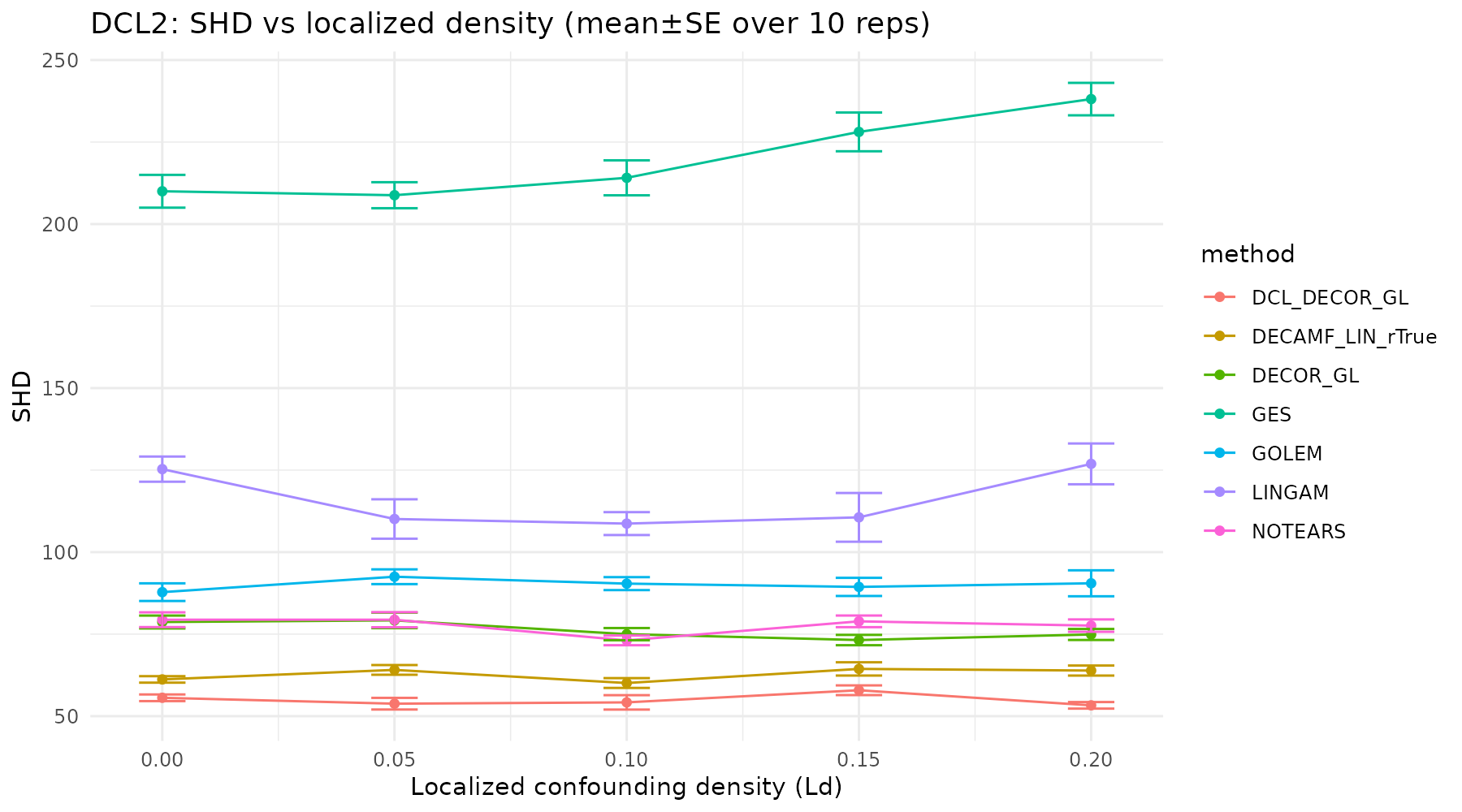}
  \end{minipage}

  \caption{Mixed-confounding synthetic experiments. (a) Mean $\Delta$F1 relative to DECOR-GL across pervasive rank ($q_P$) and strength ($U_d$). (b--c) With pervasive confounding fixed, performance as localized confounding density ($L_d$) increases (mean over 10 replicates).}
  \label{fig:dcl_synth}
\end{figure}

Figure~\ref{fig:dcl_synth}a reports the mean $\Delta$F1 relative to \textsc{DECOR-GL} across the grid.
\textsc{DCL--DECOR-GL} improves over \textsc{DECOR-GL} consistently as pervasive confounding strengthens (larger $q_P$ and/or $U_d$), matching the intended regime where separating the pervasive component before modeling localized dependence is beneficial.
Aggregated over the full DCL1 grid, \textsc{DCL--DECOR-GL} achieves higher directed-edge recovery and substantially lower SHD than \textsc{DECOR-GL} (mean F1 $0.417$ vs.\ $0.280$; mean SHD $55.3$ vs.\ $74.9$).
Methods that ignore latent structure (\textsc{NOTEARS}, \textsc{GOLEM}, \textsc{GES}, \textsc{LiNGAM}) degrade as pervasive confounding increases, typically producing denser graphs and larger SHD.
\textsc{DECAMF-LIN} shows mixed performance in this mixed regime, suggesting that explicitly modeling \emph{both} confounding types (pervasive plus localized) is important.

\paragraph{DCL2: fixing pervasive confounding and varying localized strength.}
To complement DCL1, we fix pervasive confounding at $(q_P,U_d)=(3,1.0)$ and vary the localized confounding density
$
L_d\in\{0,0.05,0.10,0.15,0.20\}
$
(10 replicates per setting).
Figure~\ref{fig:dcl_synth}b--c shows that \textsc{DCL--DECOR-GL} remains robust as localized confounding increases, while \textsc{DECOR-GL} and factor-only baselines degrade.
Averaged across DCL2, \textsc{DCL--DECOR-GL} improves both F1 and SHD relative to \textsc{DECOR-GL} (mean F1 $0.431$ vs.\ $0.266$; mean SHD $55.0$ vs.\ $76.2$), supporting our central claim that removing pervasive effects in the precision domain and then learning under correlated local noise is a strong strategy for mixed confounding.

\bibliography{refs} 

\appendix

\newpage
\appendix

\newpage
\section{Structural Properties of the D--C--L Precision Decomposition}
\label{app:structure}

In this appendix we provide a full version of Proposition~\ref{prop:lowrank-sparse-main} together with complete proofs. We use the notation from \S\ref{sec:setup}.

Recall the noise model
\[
\boldsymbol{\varepsilon} \;=\; 
\mathbf{W}\mathbf{w} \;+\; 
\mathbf{V}\mathbf{v} \;+\; 
\mathbf{U}\mathbf{u},
\]
with \(\mathbf{w}\sim\mathcal{N}(\mathbf{0},\mathbf{I}_p)\), \(\mathbf{v}\sim\mathcal{N}(\mathbf{0},\mathbf{I}_{r_S})\), \(\mathbf{u}\sim\mathcal{N}(\mathbf{0},\mathbf{I}_{r_L})\) independent, and
\[
\boldsymbol{\Omega}
=
\mathrm{Var}(\boldsymbol{\varepsilon})
=
\mathbf{W}\mathbf{W}^\top
+
\mathbf{V}\mathbf{V}^\top
+
\mathbf{U}\mathbf{U}^\top.
\]

\subsection{Noise-level components}

Let
\[
\mathbf{D}_\varepsilon
:=
(\mathbf{W}\mathbf{W}^\top)^{-1}
=
\mathrm{diag}(d_1,\ldots,d_p),
\qquad d_i>0,
\]
and define
\[
\mathbf{A}
\;:=\;
\mathbf{I}+\mathbf{V}^\top\mathbf{D}_\varepsilon\mathbf{V}
\ \in\ 
\mathbb{R}^{r_S\times r_S},
\qquad
\mathbf{C}_\varepsilon
:=
\mathbf{D}_\varepsilon\mathbf{V}\mathbf{A}^{-1}\mathbf{V}^\top\mathbf{D}_\varepsilon.
\]
The non-pervasive noise covariance and precision are
\[
\boldsymbol{\Gamma}_\varepsilon
:=
\mathbf{W}\mathbf{W}^\top+\mathbf{V}\mathbf{V}^\top,
\qquad
\mathbf{S}_\varepsilon
:=
\boldsymbol{\Gamma}_\varepsilon^{-1}
=
\mathbf{D}_\varepsilon-\mathbf{C}_\varepsilon.
\]
Writing \(\boldsymbol{\Omega}=\boldsymbol{\Gamma}_\varepsilon+\mathbf{U}\mathbf{U}^\top=\mathbf{S}_\varepsilon^{-1}+\mathbf{U}\mathbf{U}^\top\), the SMW identity gives
\begin{equation}
\label{eq:app-Omega-inv}
\boldsymbol{\Omega}^{-1}
=
\mathbf{S}_\varepsilon
-
\mathbf{S}_\varepsilon\mathbf{U}\big(\mathbf{I}+\mathbf{U}^\top\mathbf{S}_\varepsilon\mathbf{U}\big)^{-1}\mathbf{U}^\top\mathbf{S}_\varepsilon.
\end{equation}
We define the pervasive (low-rank) correction
\begin{equation}
\label{eq:app-Leps-def}
\mathbf{L}_\varepsilon
:=
\mathbf{S}_\varepsilon\mathbf{U}\big(\mathbf{I}+\mathbf{U}^\top\mathbf{S}_\varepsilon\mathbf{U}\big)^{-1}\mathbf{U}^\top\mathbf{S}_\varepsilon,
\end{equation}
so that
\(
\boldsymbol{\Omega}^{-1}=\mathbf{D}_\varepsilon-\mathbf{C}_\varepsilon-\mathbf{L}_\varepsilon
=
\mathbf{S}_\varepsilon-\mathbf{L}_\varepsilon.
\)

\subsection{Full statement and proof}

\begin{proposition}[Low-rankness of \(\mathbf{L}_\varepsilon\) and locality of \(\mathbf{S}_\varepsilon\)]
\label{prop:lowrank-sparse}
Assume the model in \S\ref{sec:setup} with \(\mathbf{T}=\mathbf{I}-\mathbf{B}\) unit-diagonal and triangular under some causal order, and define \(\mathbf{D}_\varepsilon,\mathbf{A},\mathbf{C}_\varepsilon,\mathbf{S}_\varepsilon,\mathbf{L}_\varepsilon\) as above.
\begin{enumerate}[leftmargin=1.5em]
\item[\textbf{(a)}] \textbf{Low-rankness and PSD of \(\mathbf{L}_\varepsilon\).} 
Let \(\mathbf{M}:=(\mathbf{I}+\mathbf{U}^\top\mathbf{S}_\varepsilon\mathbf{U})^{-1}\). Then
\[
\mathbf{L}_\varepsilon
=
(\mathbf{S}_\varepsilon^{1/2}\mathbf{U})\,\mathbf{M}\,(\mathbf{S}_\varepsilon^{1/2}\mathbf{U})^\top
\succeq 0,
\qquad
\rank(\mathbf{L}_\varepsilon)\le r_L.
\]
Consequently, \(\mathbf{L}_x=\mathbf{T}\mathbf{L}_\varepsilon\mathbf{T}^\top\succeq 0\) with \(\rank(\mathbf{L}_x)\le r_L\).

\item[\textbf{(b)}] \textbf{Exact locality of \(\mathbf{S}_\varepsilon\) under disjoint column supports.}
Suppose the columns of \(\mathbf{V}\) have disjoint supports
\(
S_j:=\mathrm{supp}(\mathbf{V}_{\cdot j})
\)
with \(|S_j|\le s\), and each row \(i\) belongs to at most \(c\) such supports. Then
\[
\mathbf{S}_\varepsilon
\;=\;
\mathbf{D}_\varepsilon
\;-\;
\sum_{j=1}^{r_S}
\frac{1}{A_{jj}}\,
\big(\mathbf{D}_\varepsilon\mathbf{V}_{\cdot j}\big)\big(\mathbf{D}_\varepsilon\mathbf{V}_{\cdot j}\big)^\top,
\]
and each row of \(\mathbf{S}_\varepsilon\) has at most \(c(s-1)\) off-diagonal nonzeros. Equivalently, \(\mathbf{C}_\varepsilon\) is supported on a union of cliques \(\{S_j\times S_j\}\).

\item[\textbf{(c)}] \textbf{Approximate locality under controlled leakage.}
Assume each row belongs to at most \(c\) supports and each column has \(|S_j|\le s\). Let
\[
\|\mathbf{A}^{-1}\|_{\mathrm{off},\infty}
:=
\max_{j}\sum_{k\neq j}\big|(\mathbf{A}^{-1})_{jk}\big|
\;\le\; \nu
\]
for some \(\nu\ge 0\). Then for any \(i\neq \ell\),
\[
\big|(S_\varepsilon)_{i\ell}\big|
\;\le\;
(D_\varepsilon)_{ii}(D_\varepsilon)_{\ell\ell}
\Bigg( 
\sum_{j:\,i\in S_j} \frac{|V_{ij} V_{\ell j}|}{A_{jj}}
\;+\;
\nu \sum_{\substack{j:\,i\in S_j\\ k\neq j}} |V_{ij} V_{\ell k}|
\Bigg).
\]
In particular, for any threshold \(\epsilon>0\), the number of indices \(\ell\) with \(|(S_\varepsilon)_{i\ell}|\ge \epsilon\) is at most \(\mathcal{O}(cs)\) provided \(\nu\) is small enough relative to \(\epsilon\), \(\max_{i,j}|V_{ij}|\), and \(\max_i (D_\varepsilon)_{ii}\).

\item[\textbf{(d)}] \textbf{Propagation to \(\mathbf{S}_x=\mathbf{T}\mathbf{S}_\varepsilon\mathbf{T}^\top\) under bounded degree.}
Let
\[
\deg_T^{\mathrm{row}}
:=
\max_i \mathrm{NNZ}(\mathbf{T}_{i\cdot}),
\qquad
\deg_T^{\mathrm{col}}
:=
\max_i \mathrm{NNZ}(\mathbf{T}_{\cdot i}),
\qquad
\deg_{S_\varepsilon}
:=
\max_i \mathrm{NNZ}((\mathbf{S}_\varepsilon)_{i\cdot}),
\]
where \(\mathrm{NNZ}\) counts entries whose magnitude exceeds a fixed small threshold. Then each row of \(\mathbf{S}_x=\mathbf{T}\mathbf{S}_\varepsilon\mathbf{T}^\top\) has at most
\[
\mathcal{O}\!\big(\deg_T^{\mathrm{row}}\deg_{S_\varepsilon}\deg_T^{\mathrm{col}}\big)
\]
entries above a (slightly larger) threshold. In particular, if \(\mathbf{B}\) has bounded (in+out) degree \(d\), then \(\deg_T^{\mathrm{row}},\deg_T^{\mathrm{col}}\le 1+d\), and \(\mathbf{S}_x\) inherits row-locality from \(\mathbf{S}_\varepsilon\).
\end{enumerate}
\end{proposition}

\begin{proof}
\paragraph{(a)}
From \eqref{eq:app-Omega-inv} and \eqref{eq:app-Leps-def},
\[
\mathbf{L}_\varepsilon
=
\mathbf{S}_\varepsilon\mathbf{U}\mathbf{M}\mathbf{U}^\top\mathbf{S}_\varepsilon,
\qquad
\mathbf{M}
=
(\mathbf{I}+\mathbf{U}^\top\mathbf{S}_\varepsilon\mathbf{U})^{-1}\succ 0.
\]
Factor \(\mathbf{S}_\varepsilon=\mathbf{S}_\varepsilon^{1/2}\mathbf{S}_\varepsilon^{1/2}\) to obtain
\[
\mathbf{L}_\varepsilon
=
(\mathbf{S}_\varepsilon^{1/2}\mathbf{U})\,\mathbf{M}\,(\mathbf{S}_\varepsilon^{1/2}\mathbf{U})^\top,
\]
which is positive semidefinite. Its rank is at most \(\rank(\mathbf{U})\le r_L\). Congruence by invertible \(\mathbf{T}\) preserves PSD and rank, so the same holds for \(\mathbf{L}_x=\mathbf{T}\mathbf{L}_\varepsilon\mathbf{T}^\top\).

\paragraph{(b)}
Under disjoint supports, \(\mathbf{V}^\top\mathbf{D}_\varepsilon\mathbf{V}\) is diagonal: for \(j\neq k\),
\[
(\mathbf{V}^\top\mathbf{D}_\varepsilon\mathbf{V})_{jk}
=
\sum_{i=1}^p V_{ij}(D_\varepsilon)_{ii}V_{ik}
=
0,
\]
since no row \(i\) belongs to both \(S_j\) and \(S_k\). Hence \(\mathbf{A}\) is diagonal and \(\mathbf{A}^{-1}=\mathrm{diag}(A_{11}^{-1},\ldots,A_{r_S r_S}^{-1})\). Using
\(\mathbf{S}_\varepsilon=\mathbf{D}_\varepsilon-\mathbf{D}_\varepsilon\mathbf{V}\mathbf{A}^{-1}\mathbf{V}^\top\mathbf{D}_\varepsilon\),
\[
\mathbf{S}_\varepsilon
=
\mathbf{D}_\varepsilon
-
\sum_{j=1}^{r_S}
\mathbf{D}_\varepsilon\mathbf{V}_{\cdot j}\,A_{jj}^{-1}\,\mathbf{V}_{\cdot j}^\top\mathbf{D}_\varepsilon
=
\mathbf{D}_\varepsilon
-
\sum_{j=1}^{r_S}
\frac{1}{A_{jj}}
(\mathbf{D}_\varepsilon\mathbf{V}_{\cdot j})(\mathbf{D}_\varepsilon\mathbf{V}_{\cdot j})^\top.
\]
Each rank-one term has support contained in \(S_j\times S_j\). Fixing a row \(i\), there are at most \(c\) indices \(j\) with \(i\in S_j\), and within each such support \(S_j\) row \(i\) connects to at most \(|S_j|-1\le s-1\) other indices. Therefore row \(i\) has at most \(c(s-1)\) off-diagonal nonzeros.

\paragraph{(c)}
From \(\mathbf{S}_\varepsilon=\mathbf{D}_\varepsilon-\mathbf{D}_\varepsilon\mathbf{V}\mathbf{A}^{-1}\mathbf{V}^\top\mathbf{D}_\varepsilon\), for \(i\neq \ell\),
\[
(S_\varepsilon)_{i\ell}
=
-\sum_{j,k} (D_\varepsilon)_{ii}V_{ij}\,(\mathbf{A}^{-1})_{jk}\,V_{\ell k}(D_\varepsilon)_{\ell\ell}.
\]
Split into diagonal and off-diagonal parts in \((j,k)\):
\begin{align*}
|(S_\varepsilon)_{i\ell}|
&\le
(D_\varepsilon)_{ii}(D_\varepsilon)_{\ell\ell}
\left(
\sum_j \frac{|V_{ij}V_{\ell j}|}{A_{jj}}
+
\sum_{j}\sum_{k\neq j} |V_{ij}|\cdot|(\mathbf{A}^{-1})_{jk}|\cdot|V_{\ell k}|
\right)\\
&\le
(D_\varepsilon)_{ii}(D_\varepsilon)_{\ell\ell}
\left(
\sum_{j:\,i\in S_j} \frac{|V_{ij}V_{\ell j}|}{A_{jj}}
+
\sum_{j:\,i\in S_j} \sum_{k\neq j} |V_{ij}|\cdot|(\mathbf{A}^{-1})_{jk}|\cdot|V_{\ell k}|
\right).
\end{align*}
Use \(\sum_{k\neq j}|(\mathbf{A}^{-1})_{jk}|\le \nu\) and pull out \(|V_{ij}|\) to obtain the displayed bound. The row-locality conclusion follows by the same counting argument as in part (b), combined with the requirement that \(\nu\) be small enough so that leakage terms fall below the chosen threshold except on \(\mathcal{O}(cs)\) indices.

\paragraph{(d)}
Let \(N_T(i):=\{k:\ T_{ik}\neq 0\}\) (row support of \(\mathbf{T}\)). Then
\[
(\mathbf{S}_x)_{ij}
=
(\mathbf{T}\mathbf{S}_\varepsilon\mathbf{T}^\top)_{ij}
=
\sum_{k\in N_T(i)}\ \sum_{\ell\in N_T(j)} T_{ik}\,(S_\varepsilon)_{k\ell}\,T_{j\ell}.
\]
Thus \((\mathbf{S}_x)_{ij}\) can be non-negligible only if there exist \(k\in N_T(i)\) and \(\ell\in N_T(j)\) with \((S_\varepsilon)_{k\ell}\) non-negligible. Fix row \(i\). Each \(k\in N_T(i)\) has at most \(\deg_{S_\varepsilon}\) such \(\ell\), and for each such \(\ell\), there are at most \(\deg_T^{\mathrm{col}}\) indices \(j\) with \(T_{j\ell}\neq 0\). Since \(|N_T(i)|\le \deg_T^{\mathrm{row}}\), the stated bound follows by a union argument.
\end{proof}

\subsection{A sufficient condition for controlled leakage}
\label{app:leakage}

The next result gives an interpretable sufficient condition for small \(\|\mathbf{A}^{-1}\|_{\mathrm{off},\infty}\), replacing hard ``orthogonality'' assumptions with overlap and dominance parameters.

\begin{proposition}[Controlled leakage under relaxed orthogonality]
\label{prop:leakage}
Let \(\mathbf{A}=\mathbf{I}+\mathbf{V}^\top\mathbf{D}_\varepsilon\mathbf{V}\). Define
\[
m := \max_j \big|\{k \neq j : \mathrm{supp}(\mathbf{V}_{\cdot j}) \cap \mathrm{supp}(\mathbf{V}_{\cdot k}) \neq \emptyset\}\big|,
\qquad
\eta := \max_{j\neq k}\big|\mathbf{V}_{\cdot j}^\top\mathbf{D}_\varepsilon\mathbf{V}_{\cdot k}\big|,
\qquad
\tau_{\min} := \min_j A_{jj}.
\]
Let \(\rho := m\eta/\tau_{\min}\). If \(\rho<1\), then
\[
\|\mathbf{A}^{-1}\|_{\mathrm{off},\infty}
\;\le\;
\frac{\rho}{\tau_{\min}(1-\rho)}
\;=\;
\frac{m\eta}{\tau_{\min}^2\big(1-m\eta/\tau_{\min}\big)}.
\]
\end{proposition}

\begin{proof}
Write \(\mathbf{A}=\mathbf{D}+\mathbf{E}\) where \(\mathbf{D}:=\mathrm{diag}(\mathbf{A})\) and \(\mathbf{E}:=\mathbf{A}-\mathbf{D}\) (off-diagonal part). Then
\[
\mathbf{A}^{-1}
=
(\mathbf{D}(\mathbf{I}+\mathbf{D}^{-1}\mathbf{E}))^{-1}
=
(\mathbf{I}+\mathbf{D}^{-1}\mathbf{E})^{-1}\mathbf{D}^{-1}.
\]
By definition, \(\|\mathbf{D}^{-1}\|_\infty=1/\tau_{\min}\). Moreover, for each row \(j\),
\[
\sum_{k\neq j} |(\mathbf{D}^{-1}\mathbf{E})_{jk}|
=
\frac{1}{A_{jj}}\sum_{k\neq j}|A_{jk}|
\le
\frac{1}{\tau_{\min}}\cdot m\eta
=
\rho,
\]
so \(\|\mathbf{D}^{-1}\mathbf{E}\|_\infty \le \rho < 1\). Hence the Neumann series converges:
\[
(\mathbf{I}+\mathbf{D}^{-1}\mathbf{E})^{-1}
=
\sum_{t=0}^\infty (-\mathbf{D}^{-1}\mathbf{E})^t.
\]
The off-diagonal mass comes from \(t\ge 1\), giving
\[
\|\mathbf{A}^{-1}\|_{\mathrm{off},\infty}
\le
\|\mathbf{D}^{-1}\|_\infty \sum_{t=1}^\infty \|\mathbf{D}^{-1}\mathbf{E}\|_\infty^t
\le
\frac{1}{\tau_{\min}}\cdot \frac{\rho}{1-\rho},
\]
which is the desired bound.
\end{proof}

\subsection{Structure preservation under \(\mathbf{T}\)-congruence}
\label{app:structure-pres}

The next proposition formalizes how local structure classes are propagated under \(\mathbf{T}\mathbf{M}\mathbf{T}^\top\). This supports Remark~\ref{rem:beyond-sparsity}.

\begin{proposition}[Structure preservation under \(\mathbf{T}\)-congruence]
\label{prop:T-congruence}
Let \(\mathbf{T}=\mathbf{I}-\mathbf{B}\) where \(\mathbf{B}\) encodes a DAG. Consider \(\mathbf{M}\in\mathbb{R}^{p\times p}\).
\begin{enumerate}[label=(\alph*),leftmargin=1.5em]
\item \textbf{Row-sparse case.} If \(\mathbf{M}\) is \(k\)-row-sparse and \(\mathbf{B}\) has maximum (in+out)-degree at most \(d\), then \(\mathbf{T}\mathbf{M}\mathbf{T}^\top\) is \(k(1+d)^2\)-row-sparse.

\item \textbf{Banded case.} Suppose the variables are ordered and \(\mathbf{M}\) is \(b\)-banded. If the DAG respects the ordering in the sense that \(B_{ij}\neq 0 \Rightarrow |i-j|\le d_{\mathrm{DAG}}\), then \(\mathbf{T}\mathbf{M}\mathbf{T}^\top\) is \((b+2d_{\mathrm{DAG}})\)-banded.

\item \textbf{Block-diagonal case (approximate).} Let \(\mathcal{P}=\{B_1,\ldots,B_K\}\) be a partition of \(\{1,\ldots,p\}\) and assume \(\mathbf{M}\) is block-diagonal w.r.t.\ \(\mathcal{P}\). If each block has at most \(c\) cross-block edges incident to it (in either direction) in the DAG, then each off-diagonal block of \(\mathbf{T}\mathbf{M}\mathbf{T}^\top\) has at most \(\mathcal{O}(c^2)\) nonzero entries.
\end{enumerate}
\end{proposition}

\begin{proof}
\paragraph{(a)}
Let \(N_T^{\mathrm{row}}(i):=\{k:\ T_{ik}\neq 0\}\) and \(N_T^{\mathrm{col}}(\ell):=\{j:\ T_{j\ell}\neq 0\}\). Under degree bound \(d\), we have \(|N_T^{\mathrm{row}}(i)|\le 1+d\) and \(|N_T^{\mathrm{col}}(\ell)|\le 1+d\). For fixed \(i,j\),
\[
(\mathbf{T}\mathbf{M}\mathbf{T}^\top)_{ij}
=
\sum_{k\in N_T^{\mathrm{row}}(i)}\ \sum_{\ell\in N_T^{\mathrm{row}}(j)}
T_{ik}M_{k\ell}T_{j\ell}.
\]
Fix row \(i\). There are at most \(1+d\) choices of \(k\). For each \(k\), row-sparsity of \(\mathbf{M}\) gives at most \(k\) indices \(\ell\) with \(M_{k\ell}\neq 0\). For each such \(\ell\), there are at most \(1+d\) indices \(j\) with \(T_{j\ell}\neq 0\). Thus row \(i\) has at most \((1+d)\cdot k\cdot (1+d)=k(1+d)^2\) nonzeros.

\paragraph{(b)}
If \(\mathbf{M}\) is \(b\)-banded, then \(M_{k\ell}=0\) whenever \(|k-\ell|>b\). In addition, \(T_{ik}\neq 0\) implies \(|i-k|\le d_{\mathrm{DAG}}\), and similarly \(T_{j\ell}\neq 0\) implies \(|j-\ell|\le d_{\mathrm{DAG}}\). Therefore, a nonzero contribution to \((\mathbf{T}\mathbf{M}\mathbf{T}^\top)_{ij}\) requires
\(
|i-k|\le d_{\mathrm{DAG}},\ |k-\ell|\le b,\ |j-\ell|\le d_{\mathrm{DAG}}
\),
which implies \(|i-j|\le b+2d_{\mathrm{DAG}}\). Hence \(\mathbf{T}\mathbf{M}\mathbf{T}^\top\) is \((b+2d_{\mathrm{DAG}})\)-banded.

\paragraph{(c)}
Write \(\mathbf{M}\) in \(K\times K\) block form. Since \(\mathbf{M}\) is block-diagonal, any off-block entry of \(\mathbf{T}\mathbf{M}\mathbf{T}^\top\) must arise from multiplying \(\mathbf{M}\) by cross-block nonzeros of \(\mathbf{T}\). If each block participates in at most \(c\) cross-block edges, then each block-row/column of \(\mathbf{T}\) has at most \(\mathcal{O}(c)\) nonzeros outside the diagonal block. Expanding \(\mathbf{T}\mathbf{M}\mathbf{T}^\top\) and counting the ways an off-diagonal block can be hit by left- and right-multiplication yields at most \(\mathcal{O}(c^2)\) induced nonzeros per off-diagonal block.
\end{proof}

\subsection*{Moralized-graph special case (independent errors)}

In the special case \(\mathbf{V}=\mathbf{U}=\mathbf{0}\), we have \(\mathbf{S}_\varepsilon=\mathbf{D}_\varepsilon\) diagonal and
\[
\boldsymbol{\Theta}
=
\mathbf{T}\mathbf{D}_\varepsilon\mathbf{T}^\top.
\]
Under a causal order where \(\mathbf{T}=\mathbf{I}-\mathbf{B}\) is unit-diagonal and upper-triangular, \(T_{ik}\neq 0\) iff \(k=i\) or there is a directed edge \(i\to k\). Hence \(\mathrm{supp}(\boldsymbol{\Theta})\) coincides with the moralized graph of \(\mathbf{B}\) (undirected edges plus co-parent ``marriages'') \citep{loh2014high}. When \(\mathbf{V}\neq\mathbf{0}\) but \(\mathbf{S}_\varepsilon\) is (approximately) local as in Proposition~\ref{prop:lowrank-sparse}(b)--(c), Proposition~\ref{prop:lowrank-sparse}(d) shows that \(\mathbf{S}_x=\mathbf{T}\mathbf{S}_\varepsilon\mathbf{T}^\top\) augments the moralized pattern only locally, while \(\mathbf{L}_x\) contributes a low-rank pervasive component.

\section{Proofs for \S\ref{sec:pop-identities}--\S\ref{sec:identifiability}}
\label{app:proofs}

\subsection{Proof of Proposition~\ref{prop:conditional-precision}}
\label{app:proof-cond-prec}

Recall $\boldsymbol{\varepsilon}=\mathbf{W}\mathbf{w}+\mathbf{V}\mathbf{v}+\mathbf{U}\mathbf{u}$ with
$\mathbf{w}\sim\mathcal{N}(\mathbf{0},\mathbf{I}_p)$,
$\mathbf{v}\sim\mathcal{N}(\mathbf{0},\mathbf{I}_{r_S})$,
$\mathbf{u}\sim\mathcal{N}(\mathbf{0},\mathbf{I}_{r_L})$ mutually independent.
Let
\[
\boldsymbol{\gamma}
:=
\mathbf{W}\mathbf{w}+\mathbf{V}\mathbf{v},
\qquad
\boldsymbol{\gamma}\sim\mathcal{N}(\mathbf{0},\boldsymbol{\Gamma}_\varepsilon),
\qquad
\boldsymbol{\Gamma}_\varepsilon=\mathbf{W}\mathbf{W}^\top+\mathbf{V}\mathbf{V}^\top,
\]
so that $\boldsymbol{\varepsilon}=\boldsymbol{\gamma}+\mathbf{U}\mathbf{u}$ with $\boldsymbol{\gamma}\perp\!\!\!\perp \mathbf{u}$.

From \eqref{eq:sem}, $\mathbf{x}=\mathbf{T}^{-\top}\boldsymbol{\varepsilon}$, hence
\[
\mathbf{x}
=
\mathbf{T}^{-\top}\boldsymbol{\gamma}
+
\mathbf{T}^{-\top}\mathbf{U}\mathbf{u}.
\]
Conditioning on $\mathbf{u}$, the second term is deterministic and the first term remains Gaussian with covariance
\[
\mathrm{Cov}(\mathbf{x}\mid \mathbf{u})
=
\mathrm{Cov}(\mathbf{T}^{-\top}\boldsymbol{\gamma})
=
\mathbf{T}^{-\top}\,\mathrm{Cov}(\boldsymbol{\gamma})\,\mathbf{T}^{-1}
=
\mathbf{T}^{-\top}\boldsymbol{\Gamma}_\varepsilon\mathbf{T}^{-1}.
\]
Since $\boldsymbol{\Gamma}_\varepsilon\succ 0$ and $\mathbf{T}$ is invertible, the conditional covariance is SPD and
\[
\mathrm{Cov}(\mathbf{x}\mid\mathbf{u})^{-1}
=
\mathbf{T}\boldsymbol{\Gamma}_\varepsilon^{-1}\mathbf{T}^\top
=
\mathbf{T}\mathbf{S}_\varepsilon\mathbf{T}^\top
=
\mathbf{S}_x,
\]
which completes the proof.

\subsection{Proof of Lemma \ref{lem:inv-perturb}}
\label{app:proof-inv-perturb}

Let $\widehat{\mathbf{S}}_x=\mathbf{S}_x+\mathbf{\Delta}$ with $\|\mathbf{\Delta}\|_2=\delta_{S,n}$.
Whenever $\|\mathbf{S}_x^{-1}\|_2\,\|\mathbf{\Delta}\|_2<1$, the matrix $\widehat{\mathbf{S}}_x$ is invertible and
\[
\widehat{\mathbf{S}}_x^{-1}-\mathbf{S}_x^{-1}
=
(\mathbf{S}_x+\mathbf{\Delta})^{-1}-\mathbf{S}_x^{-1}
=
-\mathbf{S}_x^{-1}\mathbf{\Delta}(\mathbf{S}_x+\mathbf{\Delta})^{-1}.
\]
Taking operator norms and using $\|(\mathbf{S}_x+\mathbf{\Delta})^{-1}\|_2\le \|\mathbf{S}_x^{-1}\|_2/(1-\|\mathbf{S}_x^{-1}\|_2\|\mathbf{\Delta}\|_2)$
(which follows from the Neumann series bound), we obtain
\[
\big\|\widehat{\mathbf{S}}_x^{-1}-\mathbf{S}_x^{-1}\big\|_2
\le
\|\mathbf{S}_x^{-1}\|_2\,\|\mathbf{\Delta}\|_2\,\|(\mathbf{S}_x+\mathbf{\Delta})^{-1}\|_2
\le
\frac{\|\mathbf{S}_x^{-1}\|_2^2\,\delta_{S,n}}{1-\|\mathbf{S}_x^{-1}\|_2\,\delta_{S,n}},
\]
which is desired bound
after substituting $\widehat{\boldsymbol{\Sigma}}_{\mathrm{cond}}=\widehat{\mathbf{S}}_x^{-1}$ and
$\boldsymbol{\Sigma}_{\mathrm{cond}}=\mathbf{S}_x^{-1}$.

\subsection{Proof of Theorem~\ref{thm:reduction}}
\label{app:proof-reduction}

Write the population precision as $\boldsymbol{\Theta}:=\boldsymbol{\Sigma}^{-1}$. In the D--C--L model we have the population decomposition
\begin{equation}
\label{eq:app-dcl-split}
\boldsymbol{\Theta}=\mathbf{S}_x-\mathbf{L}_x,
\qquad \mathbf{S}_x\succ 0,\ \mathbf{L}_x\succeq 0\ \text{(low-rank)}.
\end{equation}
Under Assumption~\ref{ass:transversality}, the structured--low-rank split is identifiable at the population level, i.e., the pair
$(\mathbf{S}_x,\mathbf{L}_x)$ is uniquely determined by $\boldsymbol{\Theta}$ (equivalently by $\boldsymbol{\Sigma}$).
In particular, $\mathbf{S}_x$ is identifiable from $\boldsymbol{\Sigma}$.

\paragraph{Step 1: Identifying $\boldsymbol{\Sigma}_{\mathrm{cond}}$.}
By Proposition~\ref{prop:conditional-precision}, the conditional covariance after removing pervasive confounding satisfies
\begin{equation}
\label{eq:app-sigcond}
\boldsymbol{\Sigma}_{\mathrm{cond}}
:=
\mathrm{Cov}(\mathbf{x}\mid \mathbf{u})
=
\mathbf{S}_x^{-1}.
\end{equation}
Because $\mathbf{S}_x$ is identifiable from $\boldsymbol{\Sigma}$, the matrix $\boldsymbol{\Sigma}_{\mathrm{cond}}=\mathbf{S}_x^{-1}$ is also identifiable as a (deterministic) function of $\boldsymbol{\Sigma}$.
Assumption $\kappa(\mathbf{S}_x)<\infty$ is not needed for this population identification but will be used to control stability under estimation in Theorem~\ref{thm:end2end}.

\paragraph{Step 2: The identifiable object from $\boldsymbol{\Sigma}_{\mathrm{cond}}$.}
Conditioning on $\mathbf{u}$ yields the correlated-noise SEM~\eqref{eq:cond-sem},
\[
\mathbf{x}^\perp=\mathbf{B}^\top \mathbf{x}^\perp+\boldsymbol{\varepsilon}^\perp,
\qquad
\boldsymbol{\varepsilon}^\perp\sim \mathcal{N}(\mathbf{0},\boldsymbol{\Gamma}_\varepsilon),
\qquad
\boldsymbol{\Sigma}_{\mathrm{cond}}=\mathbf{T}^{-\top}\boldsymbol{\Gamma}_\varepsilon\mathbf{T}^{-1},
\]
with $\mathbf{T}=\mathbf{I}-\mathbf{B}$. By the bow-free assumption, the true conditional parameters
$(\mathbf{B},\boldsymbol{\Gamma}_\varepsilon)$ belong to the bow-free model family.

Let $\mathcal{E}_{\mathrm{bow}}(\boldsymbol{\Sigma}_{\mathrm{cond}})$ be the set of \emph{all} bow-free parameter pairs
$(\mathbf{B}',\boldsymbol{\Gamma}'_\varepsilon)$ that reproduce the same conditional covariance $\boldsymbol{\Sigma}_{\mathrm{cond}}$
, and let $\mathcal{E}_{\mathrm{bow}}^{\min}(\boldsymbol{\Sigma}_{\mathrm{cond}})$ be the sparsity-minimal subset.
By construction, both sets depend on the data-generating process only through $\boldsymbol{\Sigma}_{\mathrm{cond}}$.
Since $\boldsymbol{\Sigma}_{\mathrm{cond}}$ is identifiable from $\boldsymbol{\Sigma}$ (Step~1), the sets
$\mathcal{E}_{\mathrm{bow}}(\boldsymbol{\Sigma}_{\mathrm{cond}})$ and $\mathcal{E}_{\mathrm{bow}}^{\min}(\boldsymbol{\Sigma}_{\mathrm{cond}})$ are also identifiable from $\boldsymbol{\Sigma}$.

Finally, because the true conditional SEM is bow-free, its (unknown) parameters lie in $\mathcal{E}_{\mathrm{bow}}(\boldsymbol{\Sigma}_{\mathrm{cond}})$, and hence the statistically meaningful identifiable causal target---when the graph is not assumed known a priori---is the corresponding minimal bow-free equivalence class
$\mathcal{E}_{\mathrm{bow}}^{\min}(\boldsymbol{\Sigma}_{\mathrm{cond}})$.
This proves the claim.

\subsection{Proof of Theorem~\ref{thm:end2end}}
\label{app:proof-end2end}

We prove the two conclusions in order.

\paragraph{Step 1: $\widehat{\boldsymbol{\Sigma}}_{\mathrm{cond}}\to\boldsymbol{\Sigma}_{\mathrm{cond}}$.}
By assumption, $\delta_{S,n}=\|\widehat{\mathbf{S}}_x-\mathbf{S}_x\|_2\to 0$.
Since $\mathbf{S}_x\succ 0$ and $\kappa(\mathbf{S}_x)<\infty$, we have $\lambda_{\min}(\mathbf{S}_x)>0$.
Therefore, for all sufficiently large $n$, $\delta_{S,n}<\lambda_{\min}(\mathbf{S}_x)$, so Lemma~\ref{lem:inv-perturb} applies and yields
\[
\big\|\widehat{\boldsymbol{\Sigma}}_{\mathrm{cond}}-\boldsymbol{\Sigma}_{\mathrm{cond}}\big\|_2
=
\big\|\widehat{\mathbf{S}}_x^{-1}-\mathbf{S}_x^{-1}\big\|_2
\;\le\;
\frac{\|\mathbf{S}_x^{-1}\|_2^2\ \delta_{S,n}}{1-\|\mathbf{S}_x^{-1}\|_2\,\delta_{S,n}}
\;\xrightarrow[n\to\infty]{}\;0,
\]
which proves $\widehat{\boldsymbol{\Sigma}}_{\mathrm{cond}}\to\boldsymbol{\Sigma}_{\mathrm{cond}}$ in operator norm.

\paragraph{Step 2: Stability of Stage~III transfers this to the causal target.}
Let $\mathcal{A}$ denote the (possibly set-valued) Stage~III mapping that takes a covariance input $\boldsymbol{\Sigma}$ to a bow-free output (e.g., a particular representative $(\widehat{\mathbf{B}},\widehat{\boldsymbol{\Gamma}}_\varepsilon)$ after optimization and bow reconciliation).
The stability assumption in the theorem is exactly that $\mathcal{A}$ is continuous at $\boldsymbol{\Sigma}_{\mathrm{cond}}$ with respect to the metric used to compare outputs (e.g., Frobenius distance between matrix representatives, or an appropriate distance between equivalence-class representatives).
Hence, as $\widehat{\boldsymbol{\Sigma}}_{\mathrm{cond}}\to\boldsymbol{\Sigma}_{\mathrm{cond}}$, we obtain
\[
\mathcal{A}\!\left(\widehat{\boldsymbol{\Sigma}}_{\mathrm{cond}}\right)
\;\longrightarrow\;
\mathcal{A}\!\left(\boldsymbol{\Sigma}_{\mathrm{cond}}\right).
\]
Under the bow-free assumption and the definition of the identifiability target 
, the population-level output
$\mathcal{A}(\boldsymbol{\Sigma}_{\mathrm{cond}})$ is an element of $\mathcal{E}_{\mathrm{bow}}^{\min}(\boldsymbol{\Sigma}_{\mathrm{cond}})$ (this is the object that Stage~III is designed to return, up to distributional equivalence within the bow-free family).
Therefore the Stage~III output computed from $\widehat{\boldsymbol{\Sigma}}_{\mathrm{cond}}$ converges to an element of
$\mathcal{E}_{\mathrm{bow}}^{\min}(\boldsymbol{\Sigma}_{\mathrm{cond}})$, completing the proof.

\end{document}